\documentclass{article}

\usepackage{PRIMEarxiv}

\usepackage{hyperref}
\usepackage{url}
\usepackage{graphicx}
\usepackage{booktabs}
\usepackage{longtable}
\usepackage{array}
\usepackage{xurl}
\usepackage{multirow}
\usepackage{wrapfig}
\usepackage{amsmath}

\usepackage[table,dvipsnames]{xcolor} 
\usepackage{arydshln}
\usepackage[most]{tcolorbox}
\usepackage{tabularx} 

\definecolor{gainred}{RGB}{210, 0, 0}
\definecolor{lossgreen}{RGB}{0, 140, 90}
\definecolor{tableheader}{HTML}{FFF3E3}
\definecolor{tableodd}{HTML}{FFF3EC}

\newcommand{\ModelIcon}[1]{%
  \raisebox{-.24ex}{%
    \includegraphics[height=1.55ex]{assets/model-family-icons/#1}%
  }\hspace{.38em}%
}

\newcommand{\IconModel}[2]{\csname #1Icon\endcsname\hspace{.38em}#2}

\newcommand{\inc}[1]{%
  \textcolor{gainred}{{\scriptsize\hspace{1pt}(+#1)}}}
\newcommand{\dec}[1]{%
  \textcolor{lossgreen}{{\scriptsize\hspace{1pt}(-#1)}}}
\definecolor{lightgray}{gray}{0.96} 
\definecolor{oursblue}{RGB}{232,240,250}
\definecolor{findingpink}{RGB}{254,225,221}
\definecolor{findingpinklight}{RGB}{254,244,243}
\definecolor{findingpinkframe}{RGB}{238,188,181}

\newtcolorbox{findingbox}[1]{
  enhanced,
  colback=findingpinklight,
  colframe=findingpinkframe,
  colbacktitle=findingpink,
  coltitle=black,
  boxrule=0.6pt,
  arc=3pt,
  left=5pt,
  right=5pt,
  top=4pt,
  bottom=4pt,
  toptitle=2pt,
  bottomtitle=2pt,
  title={#1},
  fonttitle=\small\bfseries,
  fontupper=\small
}

\title{Do Correct Answers Have Stable Latent Knowledge Support?}

\author{
  Chuancheng Shi\thanks{Equal contribution.
  Email for Chuancheng Shi: \texttt{cshi0459@uni.sydney.edu.au}.} \\
  University of Sydney \\
  Sydney, Australia
  \And
  Shangze Li\footnotemark[1] \\
  Nanjing University of Science and Technology \\
  Nanjing, China
  \And
  Wenhua Wu \\
  University of Sydney \\
  Sydney, Australia
  \And
  Kun Hu \\
  Edith Cowan University \\
  Perth, Australia
  \And
  Fei Shen\thanks{Corresponding author.
  Email: \texttt{shenfei29@nus.edu.sg}.} \\
  National University of Singapore \\
  Singapore
  \And
  Zhiyong Wang \\
  University of Sydney \\
  Sydney, Australia
  \And
  Tat-Seng Chua \\
  National University of Singapore \\
  Singapore
}

\begin{document}
\maketitle

\begin{abstract}
Large language models are typically evaluated in terms of the correctness of their output.
However, output correctness alone does not confirm whether the latent knowledge supporting an answer has stable internal representations. 
We therefore propose latent state stability (LSS) and knowledge topology stability (KTS) to investigate the stability of individual knowledge states and their relational structure, respectively.
Specifically, we first separately average the model's internal representations of the same knowledge under each elicitation condition. 
Using the mean representation under the baseline condition as a reference, we remove shared shifts to obtain corrected representations. 
We then formulate LSS as the distance between each corrected representation and its reference representation, normalized against the typical distance between knowledge items of the same type. 
We formulate KTS as the cross-condition consistency of the ordered pairwise representation distances among knowledge items of the same type. 
Next, we construct EpiStableBench, which comprises 4,411 items spanning factual knowledge and mathematical reasoning, with 76,766 controlled queries. 
Across 16 LLMs from four families, representation stability varies substantially even among items answered correctly under the base condition.
Within models, higher stability generally corresponds to higher accuracy under held-out elicitation conditions; across models, stability does not consistently increase with model size or overall accuracy.
Meanwhile, we present LatentDPO to learn a knowledge-editing policy that accounts for the stability of latent knowledge. 
Experimental results show that integrating LatentDPO with multiple existing editing methods consistently improves editing success and generalization across different expressions.
\end{abstract}

\section{Introduction}
\label{sec:intro}
Large language models (LLMs)~\cite{grattafiori2024llama,yang2024qwen25} rely on internal knowledge to perform factual question answering and mathematical reasoning~\cite{petroni2019language,meng2022locating}, yet existing evaluations primarily focus on observable outputs and provide limited insight into the stability of the internal knowledge representations underlying correct answers. 

\begin{figure*}[t]
\centering
\includegraphics[width=1\linewidth]{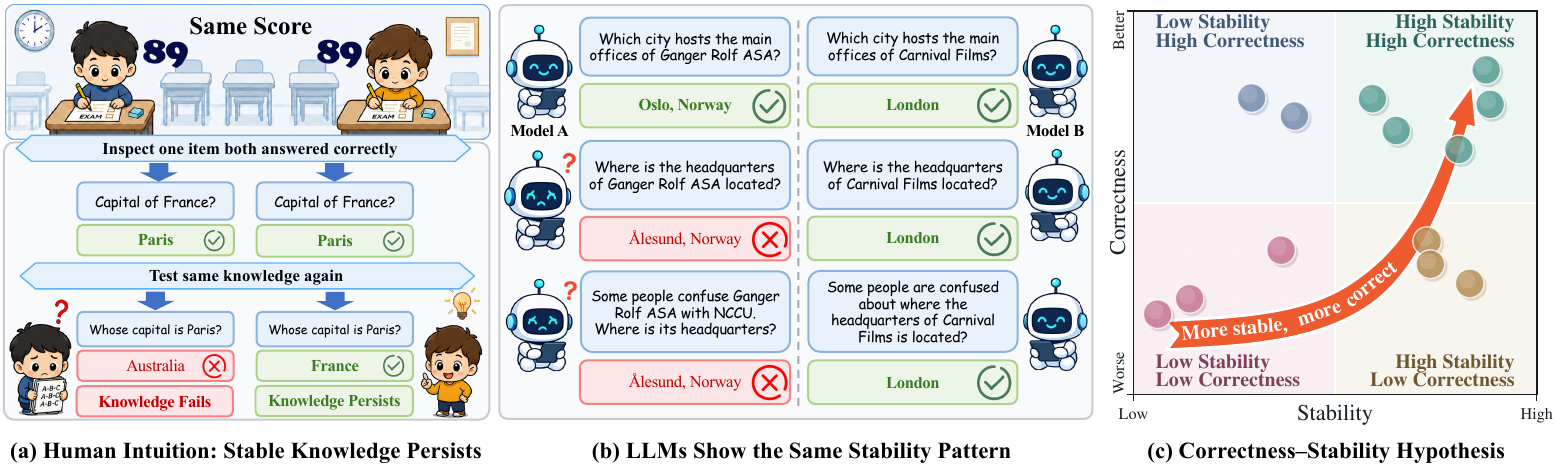}
\caption{\textbf{Motivation For The Correctness–Stability Hypothesis.}
(a) Human intuition suggests that well-established knowledge should consistently yield correct answers. (b) In LLMs, an initially correct answer may either persist or fail in subsequent elicitation conditions. (c) These observations motivate the hypothesis that model outputs may vary with representational stability.}
\vspace{-0.5cm}
\label{fig:introduction}
\end{figure*}

Recently, neuroscience studies~\cite{bellafard2024volatile,courellis2024abstract} suggest that strong behavioral performance does not necessarily imply stable internal representational organization. 
Similarly, from Figure~\ref{fig:introduction}(a), even when humans provide the same correct answer, the internal representations supporting~\cite{zhao2025we,wang2025lost} that answer may differ in stability, potentially leading to behavioral discrepancies when the same knowledge is elicited in different ways.
Given the potential instability of human knowledge representations, we examine whether correct LLM answers~\cite{lightman2024let} rely on stable latent knowledge across formulations, contexts, and languages.
Figure~\ref{fig:introduction}(b) shows that initially correct answers can diverge across elicitation conditions~\cite{zhao2025we,ifergan2025beneath,wang2025lost}, so answer correctness alone does not ensure reliable knowledge access.
Such variation may be associated with unstable internal knowledge representations.
Based on this observation, we propose the correctness-stability hypothesis, as illustrated in Figure~\ref{fig:introduction}(c): given an initially correct answer, more stable internal representations of the same knowledge are associated with a higher likelihood of remaining correct under subsequent elicitation conditions. 
Existing accuracy and behavioral consistency metrics~\cite{hendrycks2020measuring,liang2022holistic,zhao2025we} primarily evaluate output correctness and the consistency of observable outputs and therefore cannot directly characterize this internal stability.

To address this issue, we develop an internal evaluation framework that directly characterizes representational stability. We treat the hidden state immediately before answer generation as an internal observation of the latent knowledge supporting an answer when a knowledge item is elicited~\cite{orgad2025llms,gekhman2025inside,sun2025probing,gurnee2026verbalizable}, and repeatedly probe the same knowledge while varying question formulation, format, context, and language. For each question formulation, we average the representations of the same knowledge item and use the original, unmodified question as the reference representation. Since different elicitation conditions may induce representational shifts shared across knowledge items~\cite{ifergan2025beneath,wang2025lost}, we first remove these shared shifts and then measure the residual drift specific to each knowledge item. Based on this formulation, we introduce latent state stability (LSS) and knowledge topology stability (KTS) to evaluate knowledge-item identity and knowledge-organization structure, respectively. LSS measures whether knowledge-item identity remains stable across question formulations, whereas KTS measures whether the ordering of representation distances between questions of the same type remains consistent across question formulations.

To evaluate internal knowledge stability, we construct EpiStableBench, comprising 4,411 factual knowledge and mathematical reasoning targets and 76,766 controlled queries, and evaluate 16 models from four model families. Results reveals substantial variation in internal knowledge stability even among initially correct answers, with higher stability within the same model generally corresponding to higher accuracy under held-out elicitation conditions. Cross-model comparisons further show that stability does not consistently improve with model scale or external correctness. Finally, we introduce LatentDPO to learn a knowledge-editing policy that accounts for the stability of latent knowledge. Experimental results show that integrating LatentDPO with multiple existing editing methods consistently improves editing success and generalization across different expressions, with only modest reductions in specificity. We highlight the following contributions:

\begin{itemize}

\item We propose the correctness--stability hypothesis and introduce latent knowledge stability as a complementary dimension to behavioral evaluation for assessing whether correct answers remain reliable in subsequent knowledge access.

\item We propose latent state stability (LSS) and knowledge topology stability (KTS) to measure the stability of knowledge-item identity and knowledge organization, and evaluate their robustness and behavioral relevance through multiple analyses.

\item We construct and will publicly release EpiStableBench, comprising 4,411 factual-knowledge and mathematical-reasoning items and 76,766 controlled queries, with benchmark results for 16 models from four model families.

\item We propose LatentDPO, which learns a knowledge-editing policy that accounts for latent knowledge stability, and demonstrate consistent improvements across multiple editors.
\end{itemize}

\section{Related Work}

\noindent\textbf{Evaluation of Model Knowledge.}
Language models are commonly evaluated using benchmarks such as MMLU~\cite{hendrycks2020measuring} and HELM~\cite{liang2022holistic}, which infer model knowledge from externally scored responses. Recent studies~\cite{gu2025olmes,inger2025forget,hua2025flaw,bhat2026all} show that evaluation results can vary with evaluation protocols, task formulations, scoring mechanisms, and meaning-preserving input variations. Other work~\cite{zhao2025we,faghih2026same,ifergan2025beneath,wang2025lost} examines behavioral consistency in factual and cross-lingual knowledge use, with some mechanistic studies further analyzing the relationship between output inconsistency and internal representations. Overall, prior work primarily characterizes output changes across elicitation conditions and sometimes examines their internal mechanisms. We instead use LSS/KTS to test whether the internal representations supporting correct answers remain stable in both knowledge-item identity and knowledge-organization structure.

\noindent\textbf{Internal Representations and Model Behavior.}
Recent interpretability studies~\cite{shi2026tracerouter,dou2026dna,shi2026orthoeraser,shi2026culture,li2026latent,dou2026beyond,cui2026transport,cui2026llmsvlms,cui2025fine} show that a model's external behavior may not fully reflect the information encoded internally. Hidden states can encode truthfulness, error types, and correct answers that are not expressed in generation~\cite{orgad2025llms,gekhman2025inside,sun2025probing}, while the Jacobian Lens~\cite{gurnee2026verbalizable} reveals unspoken intermediate concepts that can influence subsequent outputs. These studies primarily examine what is encoded during a single inference or which internal representations causally contribute to the current output. We examine a complementary question: whether the internal knowledge representations supporting correct answers remain stable when the same knowledge is repeatedly elicited under different elicitation conditions.

\section{From Behavior to Representation Geometry}
\label{sec:method}

Output correctness does not reveal whether the latent knowledge supporting an answer remains stable
across elicitation conditions. Let
$\mathcal I=\mathcal I_{\mathrm F}\cup\mathcal I_{\mathrm M}$ denote the
factual and mathematical items, written as $k_i=(x_i,r_i,y_i)$, where
$x_i$ is the queried content, $r_i$ is its benchmark-defined comparison group,
and $y_i$ is the target answer; these correspond to
subject--relation--object and problem--task-type--result, respectively. Each
item is elicited through a reference elicitation condition $a$ and four non-reference elicitation-condition families $t\in\mathcal T$: paraphrase, format, context, and multilingual. Let
$\mathcal Q_{i,t}$ contain the target-preserving queries under condition $t$,
and let $\mathcal T_i\subseteq\mathcal T$ contain the conditions available for
item $i$. The metrics characterize internal representations across these
elicitation conditions without using generated answers or correctness labels.

\subsection{Condition-Level Representations}
\label{sec:common-space}

For model $m$, layer $\ell$, item $i$, and query $q$, let
$h^{\ell}_{m,i,q}$ denote the residual-stream state at the final input token,
immediately before generation of the first answer token. We map this state to a
model-specific common comparison space across layers using the J-Lens transport
$J^{\ell}_m$. For every available knowledge-item--elicitation-condition pair, we then average the
transported query representations to obtain a mean representation:
\begin{equation}
 z^{\ell}_{m,i,q}=J^{\ell}_m h^{\ell}_{m,i,q},
 \qquad
 \bar v^{\ell}_{m,i,t}=\frac{1}{|\mathcal Q_{i,t}|}
 \sum_{q\in\mathcal Q_{i,t}}z^{\ell}_{m,i,q}.
 \label{eq:transport-centroid}
\end{equation}
Both stability measures are defined on these centroids. Accordingly, they
characterize how the condition-level representation of an item changes across
elicitation conditions.

Condition families may induce shared shifts across knowledge items. To
separate these shared shifts from item-specific residuals, we fit a similarity map
from each non-baseline elicitation condition to the baseline elicitation condition. Let
$\tau_i\in\{\mathrm F,\mathrm M\}$ denote the track containing item $i$.
Let $\mathsf{Sim}$ be the family of maps $A(v)=sRv+b$ with $s>0$ and
$R^\top R=I$. For every model, layer, track, and non-baseline elicitation condition, we
estimate
\begin{equation}
 A^{\ell}_{m,\tau,t}
 =\arg\min_{A\in\mathsf{Sim}}
 \sum_{j\in\mathcal I^{\mathrm{tr}}_{\tau,t}}
 \left\|A(\bar v^{\ell}_{m,j,t})-\bar v^{\ell}_{m,j,a}\right\|_2^2.
 \label{eq:procrustes}
\end{equation}
All queries associated with a knowledge item remain in the same fold, and the map
applied to item $i$ is estimated without that item; for the baseline elicitation condition,
$A^{\ell}_{m,\tau,a}$ is the identity map.
The alignment removes the component of shared shifts across elicitation conditions explained by a
shared similarity transformation. The remaining variation is therefore an
item-specific residual under this alignment, rather than a uniquely identified
change in knowledge content.

\subsection{Latent State Stability}
\label{sec:iss}

Latent State Stability (LSS) measures the residual displacement of an knowledge item
from its reference representation. For item $i$ under condition $t$, define
\begin{equation}
 e^{\ell}_{m,i,t}
 =A^{\ell}_{m,\tau_i,t}(\bar v^{\ell}_{m,i,t})
 -\bar v^{\ell}_{m,i,a}.
 \label{eq:iss-residual}
\end{equation}
Because representation scales can differ across comparison groups, we
normalize this displacement by the typical anchor-space separation between
distinct knowledge items in the same group. Let $\mathcal I_r=\{j:r_j=r\}$, and let
$\beta^{\ell}_{m,r}$ be the median of
$\|\bar v^{\ell}_{m,j,a}-\bar v^{\ell}_{m,k,a}\|_2$ over unordered pairs
$j<k$ in $\mathcal I_r$. The normalized residual is
\begin{equation}
 d^{\ell}_{m,i,t}=\frac{\|e^{\ell}_{m,i,t}\|_2}
 {\beta^{\ell}_{m,r_i}+\epsilon},
 \label{eq:iss-normalized}
\end{equation}
where $\epsilon>0$ is a fixed numerical stabilizer. The scale
$\beta^{\ell}_{m,r}$ is estimated on the corresponding reference partition.
A value $d^{\ell}_{m,i,t}=1$ means that the residual displacement is equal to
the typical distance between two distinct knowledge items in the same comparison
group.

We take the median across non-baseline elicitation condition and average it over the selected
late-layer window $\mathcal W_m$:
\begin{equation}
 \operatorname{LSS}_{m,i}
 =\left(1+\frac{1}{|\mathcal W_m|}\sum_{\ell\in\mathcal W_m}
 \underset{t\in\mathcal T_i}{\operatorname{med}}
 d^{\ell}_{m,i,t}\right)^{-1}.
 \label{eq:iss}
\end{equation}
LSS lies in $(0,1]$ and decreases monotonically with normalized residual drift.
It is an item-level measure: higher LSS indicates that the corrected representation remains closer to its reference representation, relative to the local
reference scale.

\subsection{Knowledge Topology Stability}
\label{sec:kts}

Knowledge Topology Stability (KTS) evaluates whether the organization of knowledge items within a comparison group remains stable across elicitation conditions in J-Lens space. For a condition pair
$p=(t,t')$ and comparison group $r$, let
\begin{equation}
 \mathcal I_{r,p}=\{i:r_i=r,\ t,t'\in\mathcal T_i\},
 \label{eq:kts-shared-set}
\end{equation}
be the items observed under both conditions. For each
$u\in\{t,t'\}$, we center the representations using this common item set:
\begin{equation}
 \widetilde v^{\ell}_{m,i,u;p}
 =\bar v^{\ell}_{m,i,u}
 -\frac{1}{|\mathcal I_{r,p}|}
 \sum_{j\in\mathcal I_{r,p}}\bar v^{\ell}_{m,j,u}.
 \label{eq:kts-centering}
\end{equation}
For every unordered pair $i<j$ in $\mathcal I_{r,p}$, define the centered
cosine distance
\begin{equation}
 \Delta^{\ell}_{m,u,r;p}(i,j)
 =1-\cos\!\left(\widetilde v^{\ell}_{m,i,u;p},
 \widetilde v^{\ell}_{m,j,u;p}\right).
 \label{eq:kts-distance}
\end{equation}
Let $\boldsymbol\delta^{\ell}_{m,u,r;p}$ be the vectorization of these distances
over $i<j$. We compare the rank structure under the two conditions using
Spearman correlation:
\begin{equation}
 \rho^{\ell}_{m,r}(p)
 =\operatorname{Spearman}\!\left(
 \boldsymbol\delta^{\ell}_{m,t,r;p},
 \boldsymbol\delta^{\ell}_{m,t',r;p}\right).
 \label{eq:kts-pair}
\end{equation}
Centering removes translations shared within the group; cosine distance is
invariant to a common positive isotropic scale and orthogonal transformation;
and Spearman correlation is invariant to strictly monotone transformations of
the distances. Thus, KTS evaluates preservation of the rank ordering of
within-group pairwise distances, rather than equality of coordinates or
absolute distances. Here, topology refers operationally to this within-group
distance-rank structure.

Let $\mathcal P_r$ contain the condition pairs $p=(t,t')$ satisfying
$|\mathcal I_{r,p}|\geq n_{\min}$, where $n_{\min}=100$ in the primary
analysis, and for which both distance vectors are non-degenerate.
Sensitivity to the shared-item threshold is reported in
appendix~\ref{app:numerical-sensitivity}. We first average over valid
condition pairs and selected layers:
\begin{equation}
 \operatorname{KTS}_{m,r}
 =\underset{\ell\in\mathcal W_m,\,p\in\mathcal P_r}
 {\operatorname{mean}}\rho^{\ell}_{m,r}(p).
 \label{eq:kts-group}
\end{equation}
Let $\mathcal R_\tau$ be the comparison groups in track $\tau$. The overall
score first macro-averages growups within each track and then gives the two
tracks equal weight:
\begin{equation}
 \operatorname{KTS}_{m}
 =\frac{1}{2}\sum_{\tau\in\{\mathrm F,\mathrm M\}}
 \frac{1}{|\mathcal R_\tau|}
 \sum_{r\in\mathcal R_\tau}\operatorname{KTS}_{m,r}.
 \label{eq:kts}
\end{equation}
KTS lies in $[-1,1]$ and is defined at the comparison-group level. Associating
item $i$ with $\operatorname{KTS}_{m,r_i}$ supplies group-level geometric
context; it does not convert KTS into an item-level measurement.

\begin{figure*}[t]
\centering
\includegraphics[width=1\linewidth]{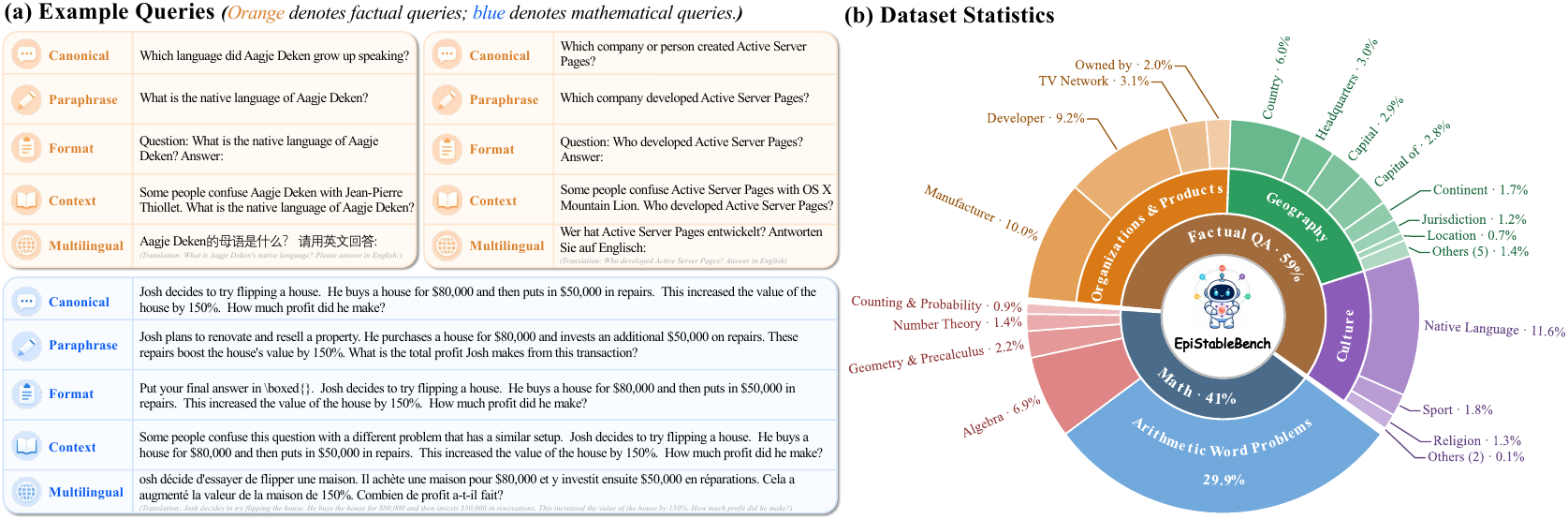}
\caption{\textbf{Overview of EpiStableBench.}
(a) Canonical questions and four controlled-query types varying paraphrase, format, context, and language for factual knowledge (\textcolor{orange}{orange}) and mathematical reasoning (\textcolor{blue}{blue}).
(b) Distribution of 4,411 knowledge items across domains and categories.}
\vspace{-0.2cm}
\label{fig:dataset}
\end{figure*}

\begin{table*}[t]
\centering
\caption{\textbf{Model-Family and Scaling Comparison.}
ACC, LSS, and KTS are reported across open-weight LLM families and parameter scales for factual knowledge, mathematical reasoning, and their average. The best and second-best results are shown in \textbf{bold} and \underline{underlined}. More results, please see the appendix.}
\label{tab:combined_leaderboard}
\resizebox{0.95\linewidth}{!}{%
\rowcolors{3}{white}{tableodd}
\begin{tabular}{c:l ccc ccc ccc}
\toprule
&
& \multicolumn{3}{c}{\textit{Factual Knowledge}}
& \multicolumn{3}{c}{\textit{Mathematical Reasoning}}
& \multicolumn{3}{c}{\textit{Average}} \\
\cmidrule(lr){3-5}
\cmidrule(lr){6-8}
\cmidrule(lr){9-11}
Family
& Model
& ACC $\uparrow$ & LSS $\uparrow$ & KTS $\uparrow$
& ACC $\uparrow$ & LSS $\uparrow$ & KTS $\uparrow$
& ACC $\uparrow$ & LSS $\uparrow$ & KTS $\uparrow$ \\
\midrule
\noalign{\vskip 1pt}

\cellcolor{white}
& \ModelIcon{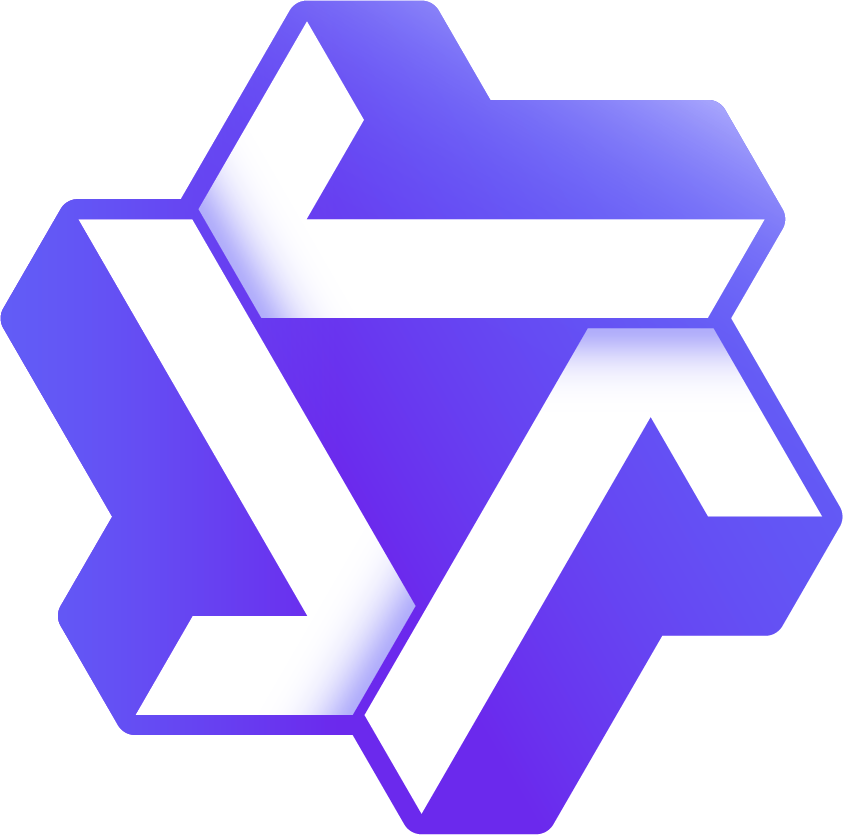}Qwen2.5-3B
& 0.766 & 0.355 & 0.795
& 0.625 & 0.596 & 0.874
& 0.696 & 0.475 & 0.834 \\

\cellcolor{white}
& \ModelIcon{qwen.png}Qwen2.5-7B
& 0.814 & 0.407 & 0.848
& 0.699 & 0.624 & 0.888
& 0.756 & 0.516 & 0.868 \\

\cellcolor{white}
& \ModelIcon{qwen.png}Qwen2.5-14B
& 0.825 & 0.445 & 0.847
& 0.744 & 0.645 & 0.892
& \underline{0.784} & 0.545 & 0.869 \\

\cellcolor{white}\multirow{-4}{*}{\textit{Qwen}}
& \ModelIcon{qwen.png}Qwen2.5-32B
& \underline{0.866} & 0.394 & 0.828
& \underline{0.756} & 0.565 & 0.881
& \textbf{0.811} & 0.480 & 0.855 \\

\midrule

\cellcolor{white}
& \ModelIcon{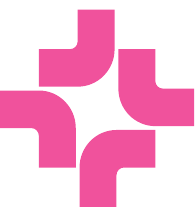}OLMo-2-1B
& 0.599 & 0.302 & 0.813
& 0.238 & 0.561 & 0.862
& 0.419 & 0.431 & 0.837 \\

\cellcolor{white}
& \ModelIcon{olmo.pdf}OLMo-2-7B
& 0.832 & 0.455 & 0.866
& 0.485 & 0.629 & 0.880
& 0.659 & 0.542 & 0.873 \\

\cellcolor{white}
& \ModelIcon{olmo.pdf}OLMo-2-13B
& 0.865 & 0.562 & \underline{0.904}
& 0.553 & 0.650 & 0.885
& 0.709 & 0.606 & 0.895 \\

\cellcolor{white}\multirow{-4}{*}{\textit{OLMo}}
& \ModelIcon{olmo.pdf}OLMo-2-32B
& \textbf{0.903} & 0.440 & 0.856
& 0.607 & 0.624 & 0.880
& 0.755 & 0.532 & 0.868 \\

\midrule

\cellcolor{white}
& \ModelIcon{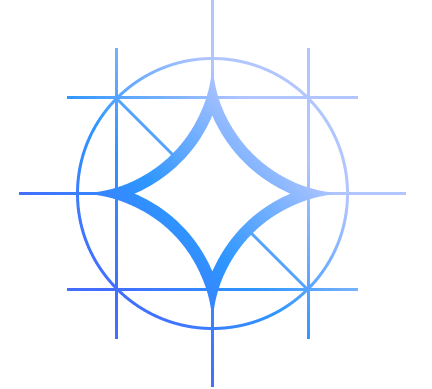}Gemma-3-1B
& 0.653 & 0.486 & 0.815
& 0.022 & 0.684 & 0.864
& 0.338 & 0.585 & 0.840 \\

\cellcolor{white}
& \ModelIcon{gemma.png}Gemma-3-4B
& 0.801 & 0.559 & 0.894
& 0.307 & \textbf{0.796} & \underline{0.914}
& 0.554 & \underline{0.677} & \underline{0.904} \\

\cellcolor{white}
& \ModelIcon{gemma.png}Gemma-3-12B
& 0.807 & 0.519 & 0.862
& 0.586 & 0.780 & \textbf{0.915}
& 0.697 & 0.650 & 0.888 \\

\cellcolor{white}\multirow{-4}{*}{\textit{Gemma}}
& \ModelIcon{gemma.png}Gemma-3-27B
& 0.808 & \underline{0.586} & 0.892
& 0.623 & \underline{0.783} & 0.911
& 0.716 & \textbf{0.684} & 0.901 \\

\midrule

\cellcolor{white}
& \ModelIcon{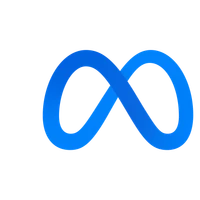}Llama3-1B
& 0.630 & 0.427 & 0.857
& 0.053 & 0.648 & 0.878
& 0.342 & 0.538 & 0.867 \\

\cellcolor{white}
& \ModelIcon{llama.png}Llama3-3B
& 0.779 & 0.546 & 0.887
& 0.218 & 0.688 & 0.883
& 0.499 & 0.617 & 0.885 \\

\cellcolor{white}
& \ModelIcon{llama.png}Llama3-8B
& 0.719 & 0.576 & 0.900
& 0.410 & 0.715 & 0.894
& 0.565 & 0.646 & 0.897 \\

\cellcolor{white}\multirow{-4}{*}{\textit{Llama3}}
& \ModelIcon{llama.png}Llama3-70B
& 0.785 & \textbf{0.638} & \textbf{0.923}
& \textbf{0.772} & 0.693 & 0.902
& 0.778 & 0.666 & \textbf{0.912} \\

\bottomrule
\end{tabular}%
}
\vspace{-0.2cm}
\end{table*}

\section{Experiments And Analysis}

\noindent\textbf{Dataset.}
EpiStableBench treats the underlying knowledge target, rather than an individual question, as the unit of evaluation. Each target has a fixed answer and is elicited through the original, unmodified question and four types of controlled queries that vary question formulation, format, context, and language. These queries preserve the target knowledge while providing repeated internal observations across different elicitation conditions. EpiStableBench contains factual knowledge and mathematical reasoning tracks constructed from established benchmarks. Figure~\ref{fig:dataset}(a) presents representative targets and their controlled queries, while Figure~\ref{fig:dataset}(b) summarizes the benchmark composition. In total, EpiStableBench comprises 4,411 factual knowledge and mathematical reasoning targets and 76,766 controlled queries.

\noindent\textbf{Protocol.}
We evaluate 16 models on EpiStableBench, covering four scales each from the Qwen2.5~\cite{yang2024qwen25}, OLMo-2~\cite{olmo2025olmo2}, Gemma-3~\cite{gemmateam2025gemma3}, and Llama~\cite{grattafiori2024llama} families.
All models use BF16, identical plain-text inputs without chat or system prompts, and one greedily decoded response per question; mathematical questions additionally use a fixed eight-shot chain-of-thought prefix.
Factual answers are independently judged by Qwen2.5-72B-Instruct~\cite{yang2024qwen25} and Mistral-Large-Instruct~\cite{mistralai2024mistrallarge2411}, while mathematical answers are assessed using deterministic equivalence checks.
LSS and KTS are aggregated over the final quarter of model layers.
Accuracy (ACC) and LSS are computed per knowledge item, with ACC weighting controlled-query types equally.
KTS is computed per relation or task type and macro-averaged across groups; response-level analyses assign each item its group's KTS.
Prediction experiments use five-fold stratified splits grouped by knowledge item.

\begin{figure*}[t]
\centering
\includegraphics[width=1\linewidth]{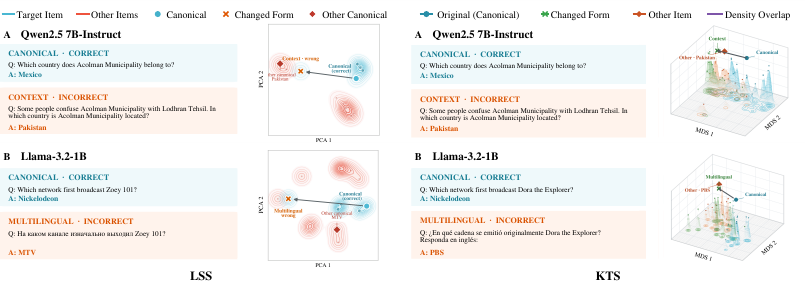}
\caption{\textbf{Behavioral Errors Accompanied By Latent Shifts.}
Items answered correctly under baseline elicitation condition produce competing answers under changed elicitation conditions, accompanied by representational shifts toward facts associated with those answers. More results please see the appendix.}
\label{fig:qualitative_shift}
\end{figure*}

\begin{figure*}[t]
\centering
\includegraphics[width=1\linewidth]{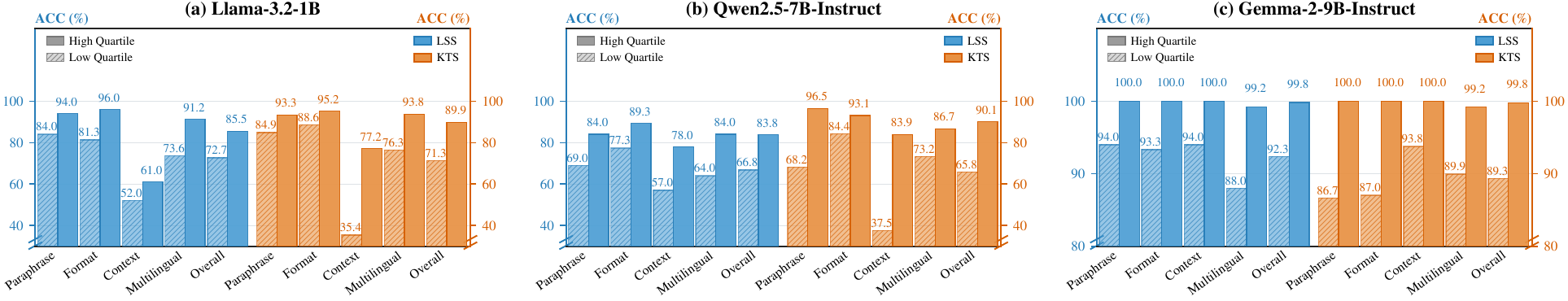}
\caption{\textbf{Stability and Subsequent Correctness.}
Among targets answered correctly under the base condition, high-LSS targets and high-KTS groups show higher ACC under held-out elicitation conditions than their low-stability counterparts.}
\label{fig:Metric_q1}
\vspace{-0.5cm}
\end{figure*}

\subsection{Model Rankings}

\noindent\textbf{Model Size, Accuracy, and Stability.}
We compare ACC, LSS, and KTS across four model families and multiple scales to assess whether larger, more accurate models have more stable knowledge representations.
Table~\ref{tab:combined_leaderboard} shows that these metrics do not consistently increase together.
From 14B to 32B, Qwen2.5's average ACC rises from 0.784 to 0.811, while LSS falls from 0.545 to 0.480 and KTS from 0.869 to 0.855; OLMo-2 shows a similar pattern from 13B to 32B.
In mathematical reasoning, ACC increases from 0.625 to 0.756 for Qwen2.5 and from 0.238 to 0.607 for OLMo-2 without corresponding stability gains.
Higher accuracy in larger models therefore does not necessarily imply greater stability of individual knowledge representations or their relational structure.

\noindent\textbf{Qualitative Analysis.}
We examine items answered correctly under the base condition but incorrectly under controlled probes (Figure~\ref{fig:qualitative_shift}).
Under context and multilingual probes, the answers for \emph{Acolman Municipality} and \emph{Zoey 101} change from Mexico to Pakistan and from Nickelodeon to MTV, respectively, accompanied by representational shifts toward competing items.
The KTS examples also exhibit changes in within-group geometry.
Together, these cases illustrate the item-level drift and relational changes captured by LSS and KTS, respectively.
Because the cases were selected after observing errors, they provide qualitative rather than predictive evidence.

\subsection{Validation of LSS and KTS}

\noindent\textbf{Stability and Subsequent Correctness.}
To test the correctness--stability hypothesis, we retain targets answered correctly under the base condition, defined by the original, unmodified question.
We hold out each controlled-query type in turn for accuracy evaluation and compute LSS and KTS from the remaining controlled queries.
Across the models and query types shown in Figure~\ref{fig:Metric_q1}, high-LSS targets and high-KTS groups consistently achieve higher ACC than their low-stability counterparts.
For Qwen2.5-7B-Instruct, ACC aggregated across query types is 83.8\% versus 66.3\% for high- and low-LSS targets, and 90.1\% versus 65.8\% for high- and low-KTS groups.
These results support the correctness-stability hypothesis: among initially correct targets, greater stability of internal knowledge representations is associated with higher accuracy under held-out elicitation conditions.

\begin{wrapfigure}{r}{0.5\textwidth}
    \centering
    \includegraphics[width=0.49\textwidth]{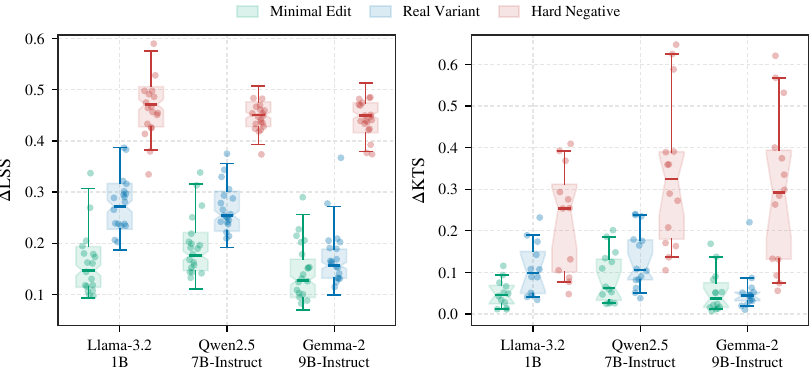}
\caption{\textbf{Semantic Sensitivity of LSS and KTS.}
Hard negatives induce larger metric changes than target-preserving perturbations across all three models.}
    \label{fig:Metric_q2}
    \vspace{-1em}
\end{wrapfigure}
\noindent\textbf{Semantic Sensitivity of Latent Knowledge Metrics.}
To assess whether LSS and KTS distinguish target knowledge changes from surface reformulations, we compare three perturbation types: \emph{minimal edits}, which add only a prefix; \emph{real variants}, which preserve the target knowledge; and \emph{hard negatives}, which replace the target fact with a similar-length fact from the same evaluation group but with a different subject and answer.
Across all three models, hard negatives produce substantially larger changes in both metrics than the two target-preserving perturbations (Figure~\ref{fig:Metric_q2}).
Changes under hard negatives are approximately 0.44--0.47 for LSS and 0.25--0.32 for KTS, compared with 0.12--0.27 and 0.03--0.11, respectively, under target-preserving perturbations.
This separation supports the semantic sensitivity of LSS and KTS, showing stronger responses to changes in target knowledge than to surface reformulations.


\begin{figure*}[h]
\centering
\includegraphics[width=1\linewidth]{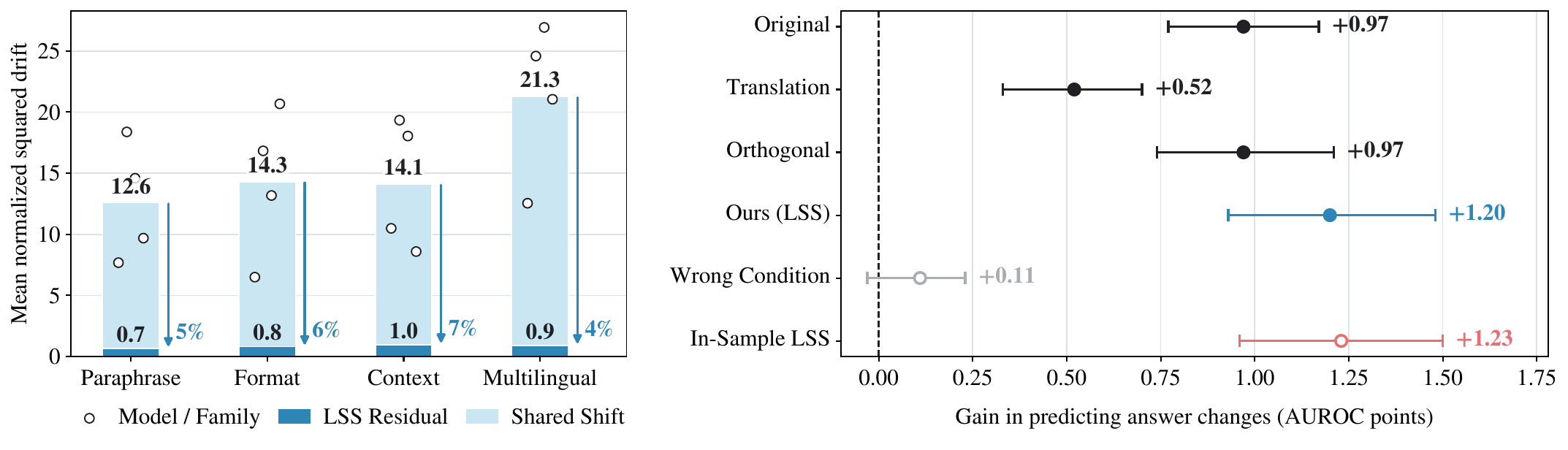}
\caption{\textbf{Separating Shared Shifts from Item-Specific Drift.}
(Left) Decomposition of normalized squared drift into the portion removed by alignment and the remaining residual.
(Right) Gains in answer-change prediction across alignment variants, measured in AUROC points.}
\label{fig:drift}
\end{figure*}

\noindent\textbf{Separating Shared Shifts from Item-Specific Drift.}
To assess whether LSS separates item-specific drift from shared shifts, we fit condition-specific similarity maps on training items and align held-out representations to the base condition.
Only 4.2\%-6.8\% of the original squared drift remains after alignment, indicating that shared condition-level transformations account for 93.2\%-95.8\% (Figure~\ref{fig:drift}(a)).
Adding the residual drift to an answer-change classifier controlling for model, relation, and condition improves AUROC by 1.20 points.
By comparison, residuals from wrong-condition alignment yield only a 0.11-point gain, with a confidence interval including zero (Figure~\ref{fig:drift}(b)).
These results suggest that condition-matched alignment reduces shared shifts while retaining item-specific variation associated with answer changes.

\begin{wraptable}{r}{0.49\textwidth}
\centering
\vspace{-1.5em}
\caption{\textbf{Relation-Aware Processing.}
Spearman correlations between ablated and full-method scores at the indicated aggregation levels. Labels before and after the slash denote LSS and KTS variants, respectively.}
\label{tab:relation_processing}
\scriptsize
\setlength{\tabcolsep}{4pt}
\renewcommand{\arraystretch}{0.98}
\begin{tabular}{lcccc}
\toprule
\multirow{2}{*}{Setting}& \multicolumn{2}{c}{\textit{LSS}} & \multicolumn{2}{c}{\textit{KTS}} \\
\cmidrule(lr){2-3}\cmidrule(lr){4-5}
& Model & Item & Model & Relation \\
\midrule
No Norm. / Centering
& 0.23 & \textbf{0.65} & \textbf{0.99} & \textbf{0.93} \\
Global Norm. / Structure
& \textbf{0.84} & 0.64 & 0.77 & 0.15 \\
\bottomrule
\end{tabular}
\vspace{-1.5em}
\end{wraptable}
\noindent\textbf{Relation-Aware Processing.}
We assess relation-aware processing by measuring rank agreement between each ablation and the full method (Table~\ref{tab:relation_processing}).
For LSS, removing normalization substantially changes model rankings, yielding a correlation of 0.23, whereas global normalization retains greater agreement at 0.84; item-level correlations are similar at 0.65 and 0.64, respectively.
For KTS, removing relation centering largely preserves model and relation rankings (0.99 and 0.93), whereas evaluating topology globally reduces these correlations to 0.77 and 0.15.
Thus, normalization affects LSS rankings, particularly across models, while relation-specific topology evaluation has a greater impact on KTS than relation centering.

\begin{wraptable}{r}{0.5\textwidth}
\centering
\vspace{-0.8em}
\caption{\textbf{Representation and Layer Selection.}
Spearman correlations with the full method, which aggregates J-Lens representations over the final 25\% of layers. Depth is normalized from input to output.}
\label{tab:representation-layers}
\scriptsize
\setlength{\tabcolsep}{4pt}
\renewcommand{\arraystretch}{0.98}
\begin{tabular}{lcccc}
\toprule
\multirow{2}{*}{Setting}& \multicolumn{2}{c}{\textit{LSS}} & \multicolumn{2}{c}{\textit{KTS}} \\
\cmidrule(lr){2-3}\cmidrule(lr){4-5}
& Model & Item & Model & Relation \\
\midrule
Depth 40\%--60\%
& 0.14 & 0.42 & -0.10 & 0.59 \\
Depth 60\%--80\%
& 0.68 & 0.77 & 0.53 & 0.85 \\
Depth 80\%--100\%
& \textbf{0.96} & \textbf{1.00} & \textbf{0.99} & \textbf{1.00} \\
Final Layer Only
& 0.60 & 0.84 & 0.98 & 0.95 \\
Raw States, Final 25\%
& 0.47 & 0.82 & 0.89 & 0.80 \\
\bottomrule
\end{tabular}
\vspace{-0.8em}
\end{wraptable}
\noindent\textbf{Representation and Layer Selection.}
We assess sensitivity to layer ranges and representation choices by comparing each variant with the full method (Table~\ref{tab:representation-layers}).
Agreement increases with depth, with the overlapping 80\%--100\% interval yielding correlations of 0.96-1.00 across all evaluation levels.
Using only the final layer or raw residual-stream states over the final 25\% of layers yields model-level LSS correlations of 0.60 and 0.47, respectively.
KTS model rankings remain more consistent under both variants, with correlations of 0.98 and 0.89.
These results indicate consistency across nearby late-layer ranges, while highlighting the sensitivity of model-level LSS rankings to aggregation and representation choices.

\begin{wraptable}{r}{0.5\textwidth}
\centering
\vspace{-1.5em}
\caption{\textbf{Probe and Reference Sensitivity.}
Spearman correlations with the full method after omitting an elicitation family or changing the reference condition.}
\label{tab:probe_reference}
\scriptsize
\setlength{\tabcolsep}{4pt}
\renewcommand{\arraystretch}{0.9}
\begin{tabular}{lcccc}
\toprule
\multirow{2}{*}{Setting}& \multicolumn{2}{c}{\textit{LSS}} & \multicolumn{2}{c}{\textit{KTS}} \\
\cmidrule(lr){2-3}\cmidrule(lr){4-5}
& Model & Item & Model & Relation \\
\midrule
\multicolumn{5}{@{}l}{\textcolor{gray}{\textit{Leave-one-probe-out}}} \\
Drop Paraphrase
& \textbf{0.98} & \textbf{0.97} & \textbf{0.99} & \textbf{0.97} \\
Drop Format
& 0.96 & 0.95 & 0.97 & 0.96 \\
Drop Context
& 0.95 & 0.95 & 0.92 & 0.91 \\
Drop Multilingual
& 0.95 & 0.95 & 0.93 & 0.90 \\
\midrule
\multicolumn{5}{@{}l}{\textcolor{gray}{\textit{Alternative Reference}}} \\
Expression Reference
& \textbf{0.96} & \textbf{0.77} & \textbf{0.98} & \textbf{0.96} \\
Format Reference
& 0.89 & 0.71 & 0.86 & 0.93 \\
\bottomrule
\end{tabular}
\vspace{-2em}
\end{wraptable}
\noindent\textbf{Probe and Reference Sensitivity.}
We assess sensitivity to elicitation-condition families and reference choices through leave-one-family-out and alternative-reference analyses (Table~\ref{tab:probe_reference}).
All correlations remain at least 0.90 when any single family is omitted, indicating consistent rankings across evaluation levels.
Using expression or format as the reference yields item-level LSS correlations of 0.77 and 0.71, respectively, while relation-level KTS correlations remain at 0.96 and 0.93.
Thus, rankings are largely stable under omission of individual condition families, whereas item-level LSS is more sensitive to the reference condition.

\subsection{Deeper Analysis}

\noindent
\begin{wrapfigure}{r}{0.5\textwidth}
\vspace{-2em}
    \centering
    \includegraphics[width=0.49\textwidth]{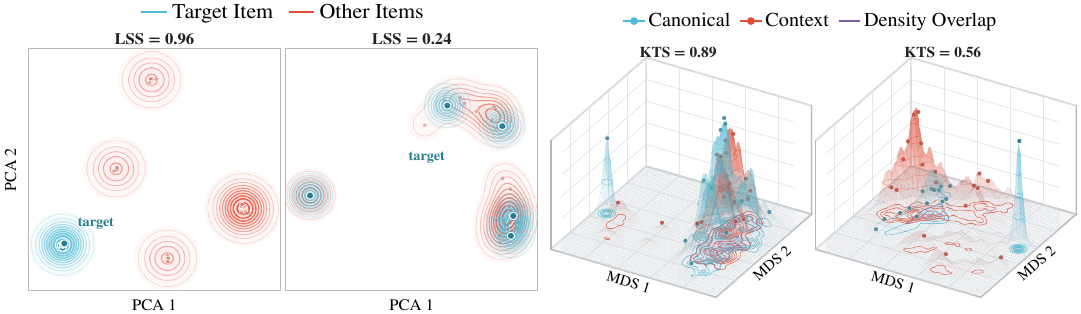}
\caption{\textbf{Geometric Patterns of Internal Knowledge Stability.}
Representative high- and low-LSS items and high- and low-KTS relations are compared across the base and controlled elicitation conditions.}
    \label{fig:combined}
    \vspace{-0.8em}
\end{wrapfigure}
\noindent\textbf{Visualizing Internal Knowledge Stability.}
We visualize representative knowledge items and relations with high and low stability (Figure~\ref{fig:combined}).
For LSS, corrected representations of high-scoring items remain close to their references, whereas those of low-scoring items exhibit greater drift toward other items.
For KTS, high-scoring relations retain more consistent within-group geometry across conditions, whereas low-scoring relations exhibit greater rearrangement.
These patterns qualitatively illustrate the item-level drift and relational changes captured by LSS and KTS, respectively.

\noindent
\begin{wrapfigure}{r}{0.5\textwidth}
    \vspace{-1em}
    \centering
    \includegraphics[width=0.9\linewidth]{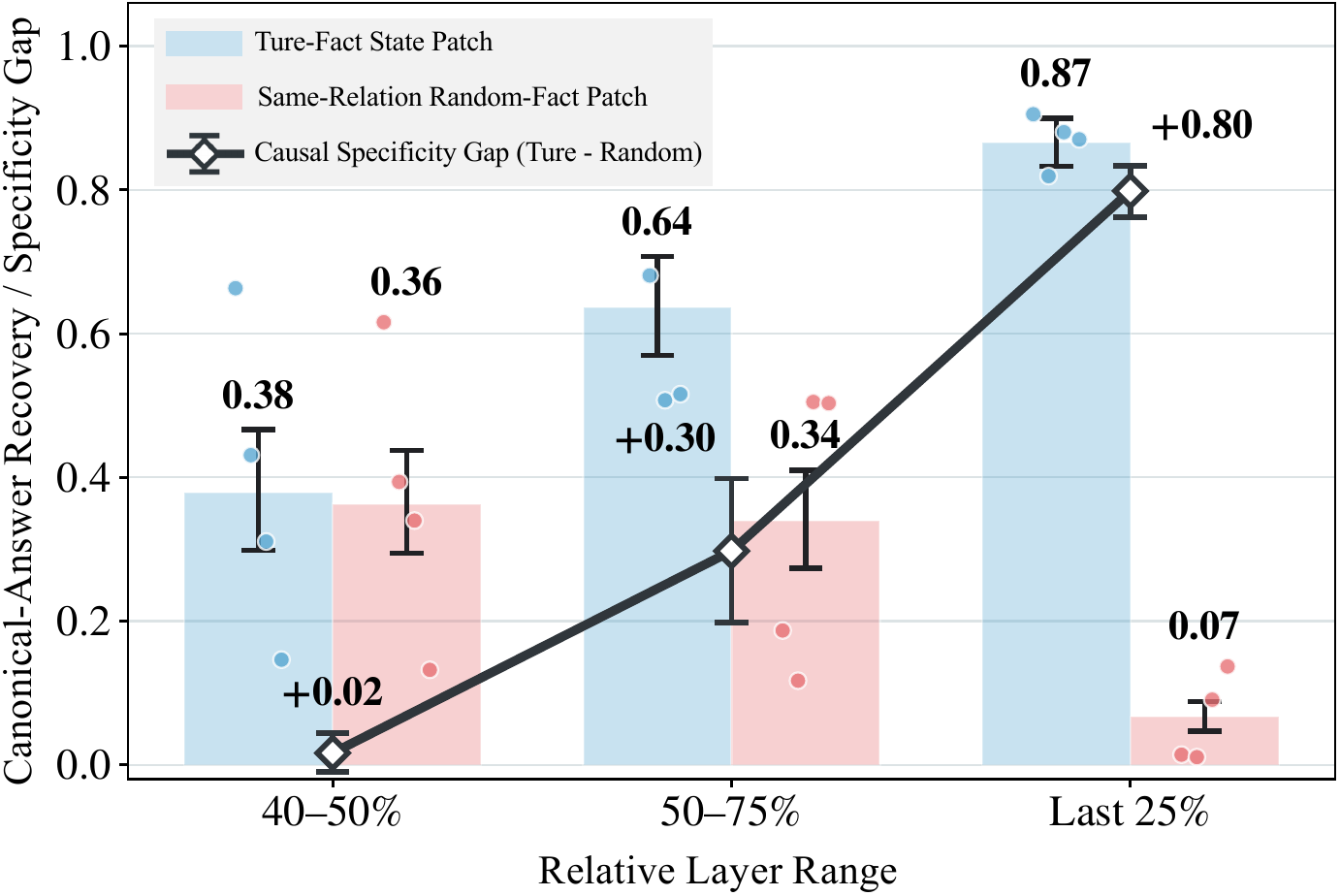}
    \vspace{-0.2cm}
\caption{\textbf{Layer-Specific Activation Patching.}
Target-answer recovery from same-item canonical states versus different-item controls within the same relation.}
    \label{fig:patch}
    \vspace{-1em}
\end{wrapfigure}
\noindent\textbf{Layer-Specific Activation Patching.}
We test whether late-layer states contain item-specific information that supports target-answer recovery.
At each tested layer, we replace a controlled query's final-input-token state with the canonical state from either the same item or a different item sharing the same relation.
The control preserves the relation type while changing item identity, helping distinguish target-specific information from signals shared within a relation.
Figure~\ref{fig:patch} shows that same-item patches increasingly outperform controls with depth.
Their respective recovery rates are 0.38/0.36 at 40\%--50\% depth, 0.64/0.34 at 50\%--75\%, and 0.87/0.07 in the final quarter, widening the recovery gap from 0.02 to 0.80.
This experiment complements the earlier correlational analyses by directly testing the functional contribution of internal states to answer recovery.
These interventions indicate that late-layer states carry target-specific information useful for answer recovery, supporting the final-quarter layer selection for LSS and KTS.

\noindent
\begin{wrapfigure}{r}{0.5\textwidth}
    \centering
    \includegraphics[width=0.49\textwidth]{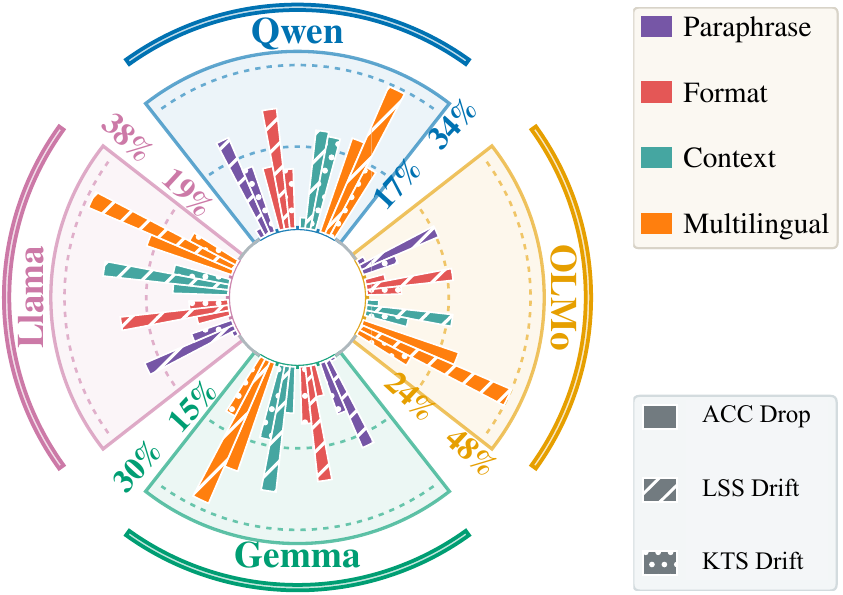}
\caption{\textbf{Family-Specific Sensitivity to Elicitation.}
Condition-wise changes in ACC, LSS, and KTS relative to the base condition across four model families.}
    \label{fig:failure_modes}
\end{wrapfigure}
\noindent\textbf{Family-Specific Sensitivity to Elicitation.}
We compare changes in ACC, LSS, and KTS relative to the base condition across model families and elicitation conditions (Figure~\ref{fig:failure_modes}).
These metrics capture output correctness, individual-state stability, and relational-structure stability, respectively.
Using a common base condition makes it possible to examine whether changes in model behavior coincide with changes in its internal representations.
The patterns vary across both families and conditions, and accuracy and stability do not always change together.
Multilingual elicitation produces large ACC drops and substantial LSS drift across multiple families, suggesting that both answer correctness and item-level representations are sensitive to language changes.
Under some other conditions, ACC changes only slightly even when LSS or KTS shows noticeable drift.
In these cases, similar output accuracy can mask changes in the states associated with individual items or in the structure of their relations.
Together, these results show why ACC alone provides an incomplete account of elicitation sensitivity: LSS and KTS reveal complementary variation at the item and relation levels.

\begin{table*}[h]
\centering
\caption{\textbf{Application to Knowledge Editing.}
Accuracy, Generality, and Specificity (\%) under identical edit sets and evaluation protocols. Parentheses report percentage-point changes relative to the baseline.}
\label{tab:policy_training_application}
\resizebox{0.95\linewidth}{!}{
\begin{tabular}{lcccccc}
\toprule
\multirow{2}{*}{Method}
& \multicolumn{3}{c}{\textit{GPT2-XL}}
& \multicolumn{3}{c}{\textit{LLaMA 2-7B}} \\
\cmidrule(lr){2-4}
\cmidrule(lr){5-7}
& Accuracy $\uparrow$
& Generality $\uparrow$
& Specificity $\uparrow$
& Accuracy $\uparrow$
& Generality $\uparrow$
& Specificity $\uparrow$ \\
\midrule

ROME
& 99.40
& 73.67
& \textbf{80.61}
& \textbf{99.90}
& 67.37
& \textbf{93.87} \\

\rowcolor{tableodd}
\textbf{ROME + Ours}
& \textbf{99.65}\inc{0.25}
& \textbf{80.48}\inc{6.81}
& 80.27\dec{0.34}
& \textbf{99.90}{\scriptsize\,(+0.00)}
& \textbf{75.75}\inc{8.38}
& 93.33\dec{0.54} \\

\midrule

MEMIT
& 72.37
& 38.89
& \textbf{87.57}
& 76.08
& 54.23
& \textbf{74.40} \\

\rowcolor{tableodd}
\textbf{MEMIT + Ours}
& \textbf{87.14}\inc{14.77}
& \textbf{54.98}\inc{16.09}
& 86.94\dec{0.63}
& \textbf{90.79}\inc{14.71}
& \textbf{69.17}\inc{14.94}
& 73.52\dec{0.88} \\

\midrule

SAKE
& 94.59
& 83.96
& \textbf{84.20}
& 96.70
& 80.73
& \textbf{84.67} \\

\rowcolor{tableodd}
\textbf{SAKE + Ours}
& \textbf{96.80}\inc{2.21}
& \textbf{91.17}\inc{7.21}
& 83.07\dec{1.13}
& \textbf{97.70}\inc{1.00}
& \textbf{88.44}\inc{7.71}
& 83.71\dec{0.96} \\
\bottomrule
\end{tabular}
}
\end{table*}

\subsection{Application}

We evaluate stability-aware policy training with ROME~\cite{meng2022locating}, MEMIT~\cite{meng2023massediting}, and SAKE~\cite{scialanga2025sake} on GPT2-XL~\cite{radford2019language} and LLaMA~2-7B~\cite{touvron2023llama2}, using matched baselines, identical edit sets, and the same evaluation protocol.
Table~\ref{tab:policy_training_application} shows improved Generality and maintained or improved Accuracy across all settings.
ROME retains near-perfect Accuracy while increasing Generality by 6.81--8.38 percentage points.
MEMIT achieves the largest gains, improving Accuracy by 14.71--14.77 points and Generality by 14.94--16.09 points.
SAKE improves Accuracy by 1.00--2.21 points and Generality by 7.21--7.71 points.
These gains accompany Specificity reductions of 0.34--1.13 points, demonstrating consistent editing benefits across editors and backbones with modest locality trade-offs.
Further implementation and evaluation details are provided in the appendix.

\section{Conclusion}
We introduce EpiStableBench and propose LSS and KTS to evaluate latent knowledge stability at the individual-state and relational-structure levels across elicitation conditions.
Experiments on 16 models from four families reveal substantial stability variation even among items answered correctly under the base condition.
Within models, higher stability generally corresponds to higher accuracy under held-out conditions; across models, stability does not consistently increase with model size or overall accuracy.
Perturbation and alignment analyses support the metrics' interpretation, while activation patching provides evidence for the functional relevance of late-layer states to target-answer recovery.
Stability-aware policy training further improves editing Generality while maintaining or improving Accuracy across three editors and two backbones, with modest reductions in Specificity.
These findings support evaluating internal representational stability alongside output correctness to better understand the knowledge supporting model answers.

\bibliographystyle{unsrt}  
\bibliography{references}  

\newpage
\appendix

\section*{Appendix}

The appendix is organized as follows. Appendix~\ref{app:notation} summarizes the notation used throughout the paper. Appendix~\ref{app:dataset-construction} describes the construction and human validation of EpiStableBench. Appendix~\ref{app:experimental-details} provides the detailed experimental setup. Appendix~\ref{More_Experiments} presents additional experiments, qualitative analyses of LSS and KTS, and quantitative comparisons across a broader set of model families. Appendix~\ref{app:application-details} details the stability-aware policy training framework and its application to knowledge editing. Appendix~\ref{app:theoretical-derivation} presents the theoretical derivation, properties, and scope of LSS and KTS. Finally, Appendices~\ref{lim} and~\ref{dis} discuss the limitations and additional questions concerning the interpretation and design choices of the proposed metrics.

\section{Notation Summary}
\label{app:notation}

Table~\ref{tab:notation} summarizes the notation used to estimate whether the
internal representations of the same knowledge remain stable across elicitation
conditions. A superscript $\ell$ indexes a model layer and does not denote
exponentiation.

\begin{table*}[h]
\centering
\caption{Core notation for Latent State Stability (LSS) and Knowledge Topology Stability (KTS).}
\label{tab:notation}
\small
\setlength{\tabcolsep}{4pt}
\renewcommand{\arraystretch}{1.00}\begin{tabular}{@{}>{\raggedright\arraybackslash}p{0.19\textwidth}p{0.77\textwidth}@{}}
\toprule
\textbf{Symbol} & \textbf{Definition} \\
\midrule
$m,\ell,i,q$ & Model, layer, knowledge item, and query, respectively. \\
$\mathcal I,\mathcal I_{\mathrm F},\mathcal I_{\mathrm M}$ & All knowledge items and the factual-knowledge and mathematical-reasoning subsets, respectively. \\
$k_i=(x_i,r_i,y_i)$ & Knowledge item $i$, where $x_i$ is the queried content, $r_i$ is its benchmark-defined comparison group, and $y_i$ is the target answer. \\
$\tau_i\in\{\mathrm F,\mathrm M\}$ & Track containing item $i$, where $\mathrm F$ and $\mathrm M$ denote the factual-knowledge and mathematical-reasoning tracks, respectively. \\
$a,\mathcal T,\mathcal T_i$ & Reference elicitation condition $a$, elicitation-condition families $\mathcal T$, and the conditions $\mathcal T_i\subseteq\mathcal T$ available for item $i$. \\
$\mathcal Q_{i,t}$ & Set of target-preserving queries for item $i$ under elicitation condition $t$. \\
$p=(t,t'),\ u$ & Pair of elicitation conditions and a condition $u\in\{t,t'\}$ within that pair. \\
$\mathcal I_r$ & Set of knowledge items in comparison group $r$, $\mathcal I_r=\{j:r_j=r\}$. \\
$\mathcal I^{\mathrm{tr}}_{\tau,t}$ & Cross-fitting training items used to estimate the similarity map for track $\tau$ and condition $t$, excluding the evaluated item. \\
$\mathcal I_{r,p}$ & Items in comparison group $r$ observed under both conditions in $p=(t,t')$. \\
$\mathcal W_m$ & Selected late-layer window for model $m$ (the final quarter of layers in the primary analysis). \\
$\mathcal P_r,\mathcal R_{\tau}$ & Valid condition pairs for comparison group $r$, and comparison groups in track $\tau$, respectively. \\
\midrule
$h^{\ell}_{m,i,q}$ & Residual-stream state at the final input token, immediately before generation of the first answer token. \\
$J^{\ell}_{m},\ z^{\ell}_{m,i,q}$ & J-Lens transport to the model-specific common comparison space, and the transported query representation. \\
$\bar v^{\ell}_{m,i,t}$ & Mean representation obtained by averaging the transported query representations in $\mathcal Q_{i,t}$. \\
$\mathsf{Sim}$ & Similarity-map family $A(v)=sRv+b$, where $s>0$ and $R^\top R=I$. \\
$A^{\ell}_{m,\tau,t}$ & Cross-fitted similarity map from elicitation condition $t$ to the reference elicitation condition $a$ for model $m$, layer $\ell$, and track $\tau$. \\
\midrule
$e^{\ell}_{m,i,t}$ & Item-specific residual between the aligned representation under condition $t$ and the reference representation under anchor $a$. \\
$\beta^{\ell}_{m,r}$ & Median reference-space distance between distinct knowledge items in comparison group $r$. \\
$\epsilon$ & Positive numerical stabilizer in the LSS normalization. \\
$d^{\ell}_{m,i,t}$ & Residual displacement normalized by the typical reference-space separation in comparison group $r_i$. \\
$D_{m,i}$ & Normalized residual drift aggregated over the available non-reference elicitation conditions and selected layers. \\
$\operatorname{LSS}_{m,i}$ & Item-level latent state stability in $(0,1]$; higher values indicate that the corrected representation remains closer to its reference representation relative to the local reference scale. \\
\midrule
$\widetilde v^{\ell}_{m,i,u;p}$ & Representation of item $i$ under condition $u\in\{t,t'\}$, centered using the common item set $\mathcal I_{r,p}$. \\
$\Delta^{\ell}_{m,u,r;p}(i,j)$ & Centered cosine distance between items $i$ and $j$ in $\mathcal I_{r,p}$. \\
$\boldsymbol\delta^{\ell}_{m,u,r;p}$ & Vector of centered cosine distances over all unordered item pairs in $\mathcal I_{r,p}$. \\
$\rho^{\ell}_{m,r}(p)$ & Spearman correlation between the distance vectors under the two conditions in $p$. \\
$\operatorname{KTS}_{m,r}$ & Comparison-group-level knowledge topology stability, averaged over valid condition pairs and selected layers. \\
$\operatorname{KTS}_{m}$ & Overall KTS, obtained by macro-averaging comparison groups within each track and assigning equal weight to the two tracks. \\
\bottomrule
\end{tabular}
\end{table*}

\section{EpiStableBench Construction}
\label{app:dataset-construction}

\noindent\textbf{Data Sources and Item Definition.}
EpiStableBench contains factual-knowledge and mathematical-reasoning tracks constructed through a shared pipeline. The factual track draws from CounterFact~\cite{meng2022locating}, LAMA~\cite{petroni2019language}, T-REx~\cite{elsahar2018t}, Wikidata~\cite{vrandevcic2014wikidata}, and PopQA~\cite{mallen2023llm_memorization}. The mathematical track draws from GSM8K~\cite{cobbe2021training} and MATH-500~\cite{lightman2024let}. We collect factual triples and mathematical problems and apply track-specific normalization and filtering to retain items with unambiguous target answers.

Both tracks represent a knowledge item as a triple $k=(s,r,o)$. For factual knowledge, $s$ is the subject, $r$ is the knowledge relation, and $o$ is the target answer. For mathematical reasoning, $s$ is the problem, $r$ is the reasoning-task type, and $o$ is the correct result. This shared representation allows factual relations and mathematical task types to serve as the corresponding evaluation groups in LSS and KTS analyses.

\noindent\textbf{Canonical Questions and Controlled Probes.}
For each knowledge item, we construct a canonical question and four types of controlled probes varying paraphrase, format, context, and multilingual presentation. All questions associated with an item preserve its target knowledge and correct answer. The controlled probes therefore provide repeated observations of how the same item is represented and used under different forms of access.

For example, the factual item ``Aagje Deken--native language--Dutch'' is represented as $k=(\text{Aagje Deken},\text{native language},\text{Dutch})$. Its canonical question and controlled probes use different presentations but continue to examine the same relation and require the answer ``Dutch.'' For a mathematical item, the presentation may change while the underlying problem, reasoning target, and correct result remain fixed.

\noindent\textbf{Benchmark Composition.}
The resulting benchmark contains 2,592 factual items from 21 relations and 1,819 mathematical problems. Across the two tracks, EpiStableBench contains 4,411 knowledge items and 76,766 queries. All evaluated models receive the same items, canonical questions, and controlled probes.

\noindent\textbf{Probe Generation and Human Validation.}
Multilingual probes were generated using ChatGPT-5.6 Sol. To ensure semantic preservation in EpiStableBench, 20 human reviewers manually screened the dataset before evaluation, each reviewing a separate subset. Reviewers checked whether rewritten questions preserved the original relation, task requirements, and intended answer. Questions containing ambiguity, semantic drift, answer changes, or other inconsistencies were revised or removed to reduce potential confounds in the evaluation of knowledge stability.

\section{Detailed Experimental Setup}
\label{app:experimental-details}

\noindent\textbf{Model Checkpoints and Hardware.}
We evaluate 16 LLMs in the main text. Qwen2.5~\cite{qwen2025qwen25technicalreport} at 3B, 7B, 14B, and 32B; OLMo-2~\cite{olmo20242} at 1B, 7B, 13B, and 32B; Gemma-3~\cite{team2025gemma} at 1B, 4B, 12B, and 27B; and Llama~\cite{grattafiori2024llama} at 1B, 3B, 8B, and 70B. All checkpoints are loaded in BF16 without quantization. Experiments are conducted on four NVIDIA H200 GPUs.

\noindent\textbf{Evaluation Data.}
EpiStableBench contains 4,411 knowledge items and 76,766 queries. The factual track contains 2,592 items from 21 relations, and the mathematical track contains 1,819 items. Each item is elicited through canonical queries and controlled queries varying paraphrase, format, context, or language. All models are evaluated on the same items and queries.

\noindent\textbf{Input and Generation Configuration.}
All models receive the same task-specific plain-text inputs without chat templates or system messages. Factual queries request the shortest correct answer, whereas mathematical queries use a fixed eight-shot chain-of-thought prefix and request a step-by-step solution. Each question is answered once using greedy decoding with \texttt{do\_sample=False} and one beam. The maximum generation length is 24 tokens for factual queries, 512 tokens for mathematical queries, and 640 tokens for multilingual mathematical queries.

\noindent\textbf{Latent-State Extraction.}
For each question and selected layer, we extract the residual-stream state at the final non-padding input token after the complete question has been processed and before the first answer token is generated. Layer depth is normalized as \(d_\ell=\ell/(L-1)\). Primary results are aggregated over the final-quarter layers with \(d_\ell\geq0.75\), while layers with \(0.50\leq d_\ell<0.75\) are examined in the sensitivity analysis.

\noindent\textbf{J-Lens and Cross-Fitted Alignment.}
J-Lens~\cite{gurnee2026verbalizable} estimates a rank-256 transport from each selected layer to the final representation space using randomized range finding. The transport is calibrated on 128 length-128 sequences sampled from \texttt{NeelNanda/pile-10k}. LSS alignment uses five-fold cross-fitting stratified by factual relation or mathematical task type. For an item $f$ in fold $\mathcal F_j$, the Procrustes parameters used to compute its residual drift are estimated only from items outside $\mathcal F_j$. All queries associated with the same knowledge item remain in the same fold.

\noindent\textbf{Evaluation Groups and KTS Notation.}
Throughout the metric definitions, $r_f$ denotes the evaluation group of item $f$: a factual relation for factual items and a task type for mathematical items. KTS is defined at this group level rather than at the individual-item level. In analyses relating KTS to item-level behavior, each item is associated with the KTS of its corresponding group.

For KTS, relation centering is performed separately for every condition pair $p=(t,t')$. Accordingly, the centered representation and pairwise distance are formally written as
\begin{equation}
\tilde v^\ell_{m,f,u;p}
=
\bar v^\ell_{m,f,u}
-
\frac{1}{|\mathcal D_{r,p}|}
\sum_{g\in\mathcal D_{r,p}}
\bar v^\ell_{m,g,u},
\qquad u\in\{t,t'\},
\end{equation}
and
\begin{equation}
\Delta^\ell_{m,u,r;p}(f,g)
=
1-\cos\!\left(
\tilde v^\ell_{m,f,u;p},
\tilde v^\ell_{m,g,u;p}
\right).
\end{equation}
Here $\mathcal D_{r,p}$ contains the items in group $r$ available under both conditions in $p$. The condition-pair index $p$ is suppressed in the main-text notation for readability.

\noindent\textbf{Answer Evaluation.}
Factual answers are evaluated independently using Qwen2.5-72B-Instruct~\cite{qwen2025qwen25technicalreport} and Mistral-Large-Instruct~\cite{mistralai2024mistrallarge}. Both judges receive the generated response, target answer, and accepted aliases and determine whether the response expresses the target answer. Mathematical answers are evaluated by extracting the final answer and applying normalized string matching, numerical equivalence with a relative tolerance of \(10^{-6}\), and symbolic equivalence with SymPy~\cite{meurer}. Outputs not classified as correct are counted as incorrect in ACC. Disagreements between the two LLM judges are very rare. We assess sensitivity using both a strict AND rule, requiring both judges to label a response as correct, and an OR rule, requiring at least one judge to do so. All disagreement cases are independently reviewed by three human annotators, blinded to the evaluated model, LSS/KTS scores, and both judges' decisions. The human majority vote determines the final adjudicated label.

\noindent\textbf{Aggregation and Statistical Analysis.}
The knowledge item is the statistical unit for ACC and LSS, whereas the factual relation or mathematical task type is the statistical unit for KTS. ACC is first averaged within each available probe type and then equally across probe types for each item. LSS is averaged across items within each track, while KTS is macro-averaged across evaluation groups within each track.

Let $\mathcal C_m$ denote the available tracks, $\mathcal F_{m,c}$ the items in track $c$, and $\mathcal R_{m,c}$ its relations or task types. Track- and model-level scores are defined as
\begin{equation}
\mathrm{LSS}_{m,c}
=
\frac{1}{|\mathcal F_{m,c}|}
\sum_{f\in\mathcal F_{m,c}}
\mathrm{LSS}_{m,f},
\qquad
\mathrm{KTS}_{m,c}
=
\frac{1}{|\mathcal R_{m,c}|}
\sum_{r\in\mathcal R_{m,c}}
\mathrm{KTS}_{m,r},
\end{equation}
\begin{equation}
\mathrm{LSS}_{m}
=
\frac{1}{|\mathcal C_m|}
\sum_{c\in\mathcal C_m}
\mathrm{LSS}_{m,c},
\qquad
\mathrm{KTS}_{m}
=
\frac{1}{|\mathcal C_m|}
\sum_{c\in\mathcal C_m}
\mathrm{KTS}_{m,c}.
\end{equation}
Thus, factual knowledge and mathematical reasoning receive equal weight in model-level comparisons. In item-level analyses, $\mathrm{KTS}_{m,r_f}$ is used as a group-level structural feature and is not interpreted as an item-specific KTS score.

We construct 95\% percentile confidence intervals using 1,000 relation- or task-type-stratified bootstrap replicates. Quartile comparisons are performed over items for LSS and over relations or task types for KTS. Predictive experiments use five-fold stratified, item-grouped splits, ensuring that observations of the same knowledge item never appear in both training and test sets.

\noindent\textbf{Activation Patching for Layer Selection.}
To examine whether late-layer representations contain more item-specific knowledge, we conduct an activation-patching experiment. For each controlled probe, we replace the native residual-stream state at the final input-token position of a decoder block with the corresponding state from the same item’s canonical question, using a different same-relation item as a control. Each run patches one block independently, once during prefill before answer generation, across sampled layers from approximately 40\% depth to the final layer. The results show that same-item patching becomes more effective in later layers, while control patching weakens. Specifically, the recovery-rate gap increases from 0.02 in the 40\%–50\% depth interval to 0.80 in the final quarter. Therefore, later-layer states more selectively support recovery of the target answer, providing intervention-based evidence for using final-quarter layers in LSS and KTS.

\section{More Experiments}
\label{More_Experiments}

\subsection{Sensitivity to Centroid Cancellation}
\label{app:centroid-cancellation}

Because condition-level centroids could in principle obscure large but opposing deviations across individual queries, we examined this possibility among items answered correctly under the reference elicitation condition. Using an independent reference panel calibrated separately by comparison group and condition, we found that severe masking events, in which individual-query representations deviated substantially while the corresponding centroid remained stable, were rare. Such events accounted for 0.289\% of query-condition observations for Qwen2.5-3B (95\% CI: 0.141\%--0.462\%) and 0.116\% for Qwen2.5-7B (95\% CI: 0.032\%--0.253\%); at the item level, 1.155\% and 0.463\% of items, respectively, exhibited such an event under at least one condition. Under a stricter criterion requiring the centroid deviation to fall below the 25th percentile of its reference distribution, the corresponding rates decreased to 0.030\% for Qwen2.5-3B and zero observed cases for Qwen2.5-7B. These results indicate that severe centroid cancellation is uncommon and does not constitute a dominant source of distortion in the centroid-based metrics. Nevertheless, centroid and tail behavior are not interchangeable: among observations whose individual-query deviations entered the top decile, 34.9\% for Qwen2.5-3B and 21.7\% for Qwen2.5-7B did not enter the top decile according to centroid deviation. We therefore additionally report tail-based results to characterize variation not captured by condition-level centroids.

\subsection{Comparison with Standard Representation Measures}
To examine whether LSS and KTS provide predictive value beyond output features, we compare them with standard representation measures. Table~\ref{tab:representation_baselines} shows that adding either metric improves failure-prediction AUROC under held-out access conditions, with their combination achieving the highest scores. Specifically, LSS/KTS reaches 0.6154 and 0.7333 on 3B and 7B, exceeding the strong output baseline by 1.61 and 1.83 percentage points and within-relation CKA by 1.21 and 1.67 points, respectively. It also outperforms the combined standard features. Therefore, these results support the complementary predictive value of internal stability, with joint item- and relation-level measurements yielding the strongest performance.

\begin{table}[h]
\centering
\small
\caption{\textbf{Comparison With Standard Representation Measures.}
All representation features augment the same strong output baseline.
Values report failure-prediction AUROC.}
\label{tab:representation_baselines}
\setlength{\tabcolsep}{5pt}
\begin{tabular}{@{}lcc@{}}
\toprule
Predictor & Qwen2.5-3B & Qwen2.5-7B \\
\midrule
Strong Output Baseline & 0.5993 & 0.7150 \\
$+$ Within-relation CKA & 0.6033 & 0.7166 \\
$+$ All Standard Features & 0.6030 & 0.7137 \\
\midrule
$+$ LSS & 0.6102 & 0.7169 \\
$+$ KTS & 0.6149 & 0.7262 \\
$+$ LSS/KTS & \textbf{0.6154} & \textbf{0.7333} \\
\bottomrule
\end{tabular}
\end{table}

\subsection{Sensitivity to J-Lens Construction and Representation Extraction}

To examine whether LSS depends on specific implementation choices, we conduct ablations over the J-Lens calibration corpus, transport rank, representation space, and token extraction position. Table~\ref{tab:iss_representation_sensitivity} shows that LSS is highly robust to choices within the J-Lens construction, while remaining specific to its defined representation space and extraction position. Specifically, changing the calibration corpus yields rank correlations of $0.964$--$0.990$. Varying the J-Lens transport rank among $\{64, 128, 256, 512\}$ yields correlations of $0.903$--$0.991$. Replacing J-Lens with raw or PCA-projected representations and replacing the final-input-token state with question-level pooling result in lower correlations of $0.373$--$0.780$ and $0.161$--$0.561$, respectively. Therefore, the LSS results are not driven by a particular J-Lens calibration corpus or transport rank, while J-Lens alignment and the pre-specified final-input-token state constitute important components of the metric definition.

\begin{table}[h]
\centering
\caption{\textbf{Sensitivity to J-Lens and Representation Choices.}
LSS rank correlations under changes to the calibration corpus,
transport rank, representation space, and token extraction position.}
\label{tab:iss_representation_sensitivity}
\begin{tabular}{lc}
\toprule
\textbf{Ablation} & \textbf{LSS Rank Correlation} \\
\midrule
Calibration Corpus      & $0.964$--$0.990$ \\
J-Lens Transport Rank   & $0.903$--$0.991$ \\
Raw/PCA Representations & $0.373$--$0.780$ \\
Question-level Pooling  & $0.161$--$0.561$ \\
\bottomrule
\end{tabular}
\vspace{-0.3cm}
\end{table}

\subsection{Reliability of KTS Across Relation Sizes}

\begin{wrapfigure}{r}{0.48\textwidth}
    \vspace{-0.8em}
    \centering
    \includegraphics[width=0.47\textwidth]{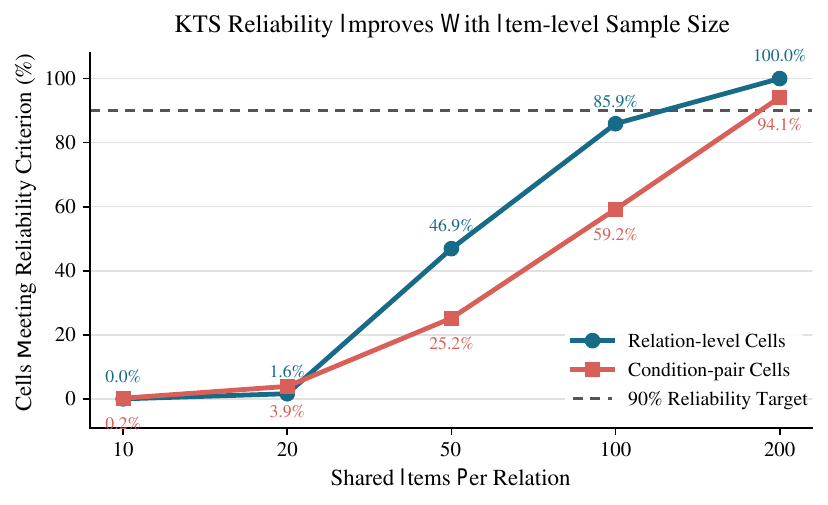}
    \vspace{-1.0em}
    \caption{\textbf{KTS Reliability Across Relation Sizes.} Reliability increases with the number of shared items, with both relation-level and condition-pair estimates exceeding the 90\% target at 200 items.}
    \label{fig:kts_reliability_rates}
    \vspace{-0.8em}
\end{wrapfigure}
To examine whether KTS is sensitive to relation size and small-sample effects, we subsample $10$, $20$, $50$, $100$, and $200$ shared items from each relation and recompute KTS at both the relation and condition-pair levels. Figure~\ref{fig:kts_reliability_rates} shows that KTS reliability improves consistently with the number of shared items. Specifically, the proportion of relation-level cells satisfying the reliability criterion increases from $0.0\%$ and $1.6\%$ at $10$ and $20$ items to $46.9\%$, $85.9\%$, and $100.0\%$ at $50$, $100$, and $200$ items, respectively. Condition-pair estimates require more observations, increasing from $0.2\%$ and $3.9\%$ to $25.2\%$, $59.2\%$, and $94.1\%$ over the same sample sizes. Therefore, approximately $100$ shared items are sufficient for exploratory relation-level analysis, whereas $200$ shared items provide a more reliable basis for fine-grained condition-pair comparisons, with both levels reaching or approaching the $90\%$ reliability target.

\begin{table}[h]
\centering
\small
\caption{\textbf{Sensitivity to Numerical and Aggregation Choices.}
Spearman correlations and maximum rank shifts are measured against
the default settings across 16 models in the audited factual cohort.
The LSS constant row summarizes all seven tested settings;
other rows concern the full method.}
\label{tab:numerical_sensitivity}
\setlength{\tabcolsep}{4pt}

\begin{tabular}{@{}clcc@{}}
\toprule
Metric & Setting Change & Rank $\rho$ & Max. Shift \\
\midrule

\multirow{3}{*}{KTS}
& Alternative Tie Conventions
& 1.000 & 0 \\

& Shared-item Threshold: $100 \to 10$
& 0.985 & 2 \\

& Item-weighted Relations
& 0.974 & 2 \\

\midrule

\multirow{2}{*}{LSS}
& Additive Constant: $0$--$10^{-3}$
& 1.000 & 0 \\

& Item-weighted Relations
& 0.988 & 2 \\

\bottomrule
\end{tabular}

\vspace{-0.3cm}
\end{table}

\subsection{Sensitivity to Numerical and Aggregation Choices}
\label{app:numerical-sensitivity}
To examine whether model rankings depend on numerical and aggregation choices, we conduct sensitivity analyses of tie handling, shared-item thresholds, the additive LSS denominator constant, and relation weighting. Table~\ref{tab:numerical_sensitivity} shows that rankings remain broadly stable within the audited factual cohort. Specifically, alternative tie conventions leave full-method KTS rankings unchanged; increasing the minimum shared-item threshold from 100 to 10 preserves a rank correlation of 0.985; LSS constants from 0 to \(10^{-3}\) leave rankings unchanged across all tested settings; and item-weighted aggregation yields rank correlations of 0.988 and 0.974 for full-method LSS and KTS, respectively. Therefore, these results support ranking robustness to the tested numerical, eligibility, and weighting choices.

\begin{figure*}[t]
\centering
\includegraphics[width=0.9\linewidth]{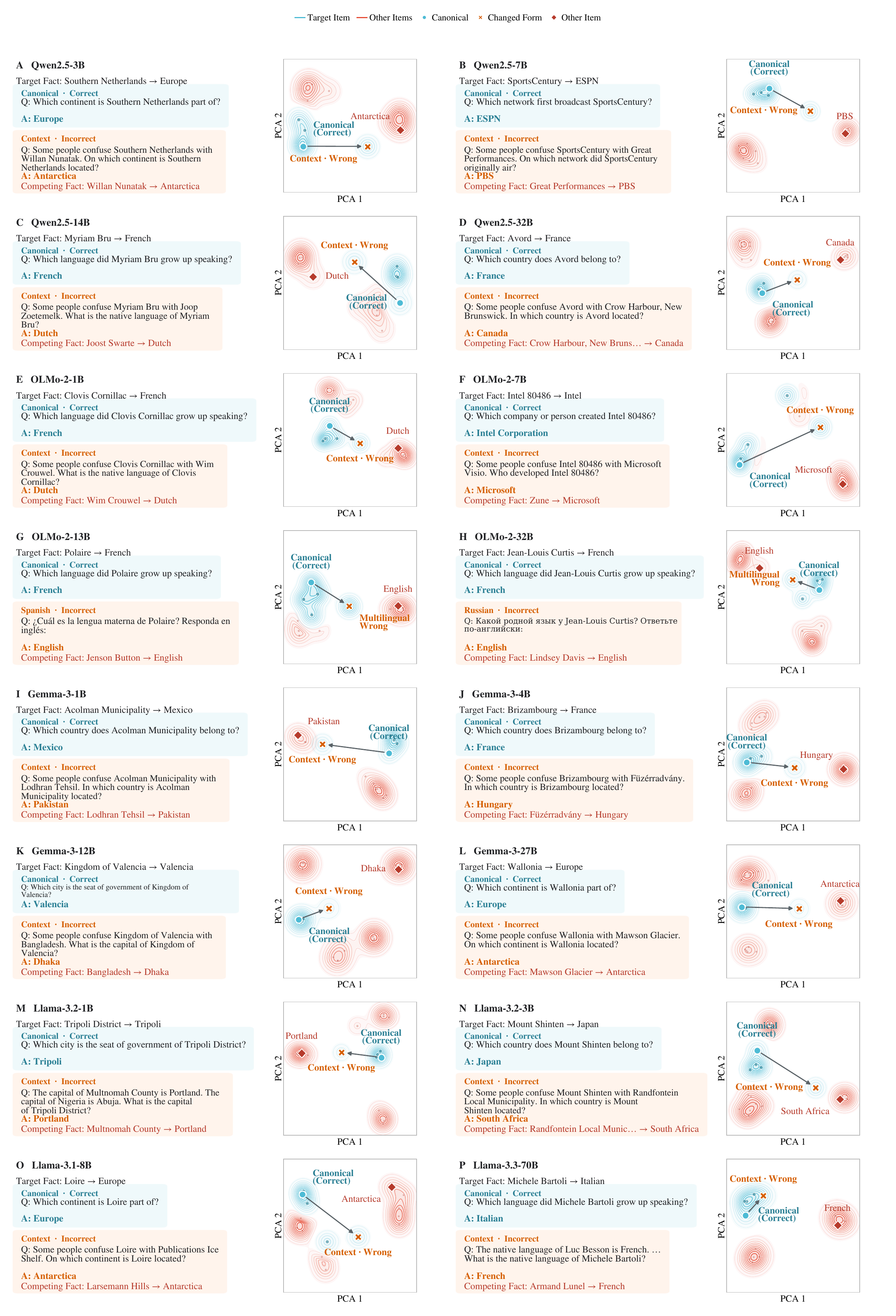}
\caption{\textbf{More Qualitative Analysis of LSS.}}
\vspace{-0.5cm}
\label{fig:more1}
\end{figure*}

\begin{figure*}[t]
\centering
\includegraphics[width=0.9\linewidth]{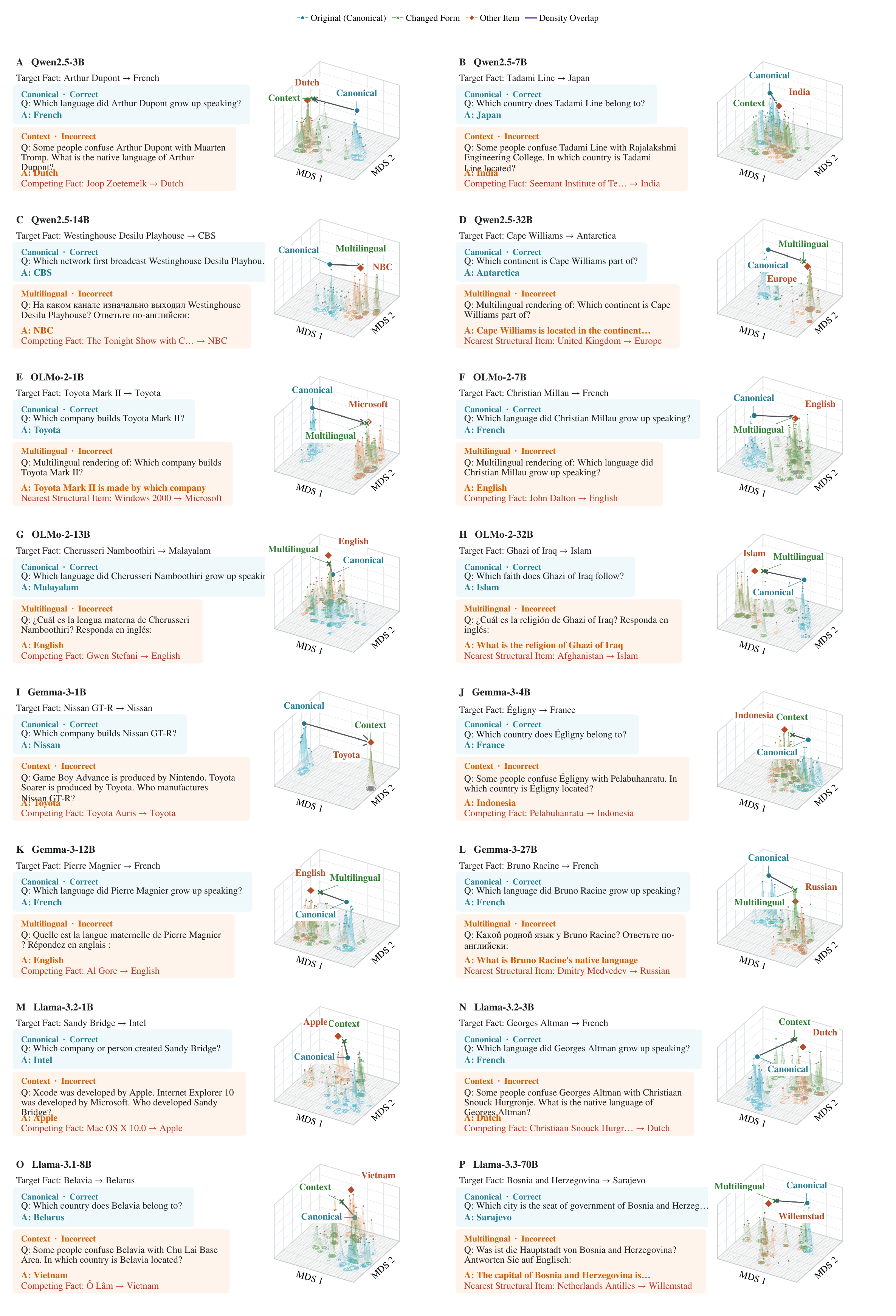}
\caption{\textbf{More Qualitative Analysis of KTS.}}
\vspace{-0.5cm}
\label{fig:more2}
\end{figure*}

\subsection{More Qualitative Analysis of LSS}
\label{more_lss}

Figure~\ref{fig:more1} extends the qualitative analysis of LSS across all 16 models. Each selected item is answered correctly under the canonical elicitation condition but incorrectly under a context or multilingual condition. Across these cases, the changed-form representation moves away from the canonical target representation and toward a region occupied by items associated with the competing answer. This recurring pattern illustrates the disruption of knowledge-item identity captured by LSS and shows that such item-specific drift occurs across model families and parameter scales. Because the examples are selected after observing answer changes, they provide qualitative rather than predictive evidence.

\subsection{More Qualitative Analysis of KTS}
\label{more_kts}

Figure~\ref{fig:more2} provides a corresponding relation-level analysis of KTS across all 16 models. Under the changed elicitation condition, the representation distribution is often displaced or mixed relative to its canonical counterpart, with reduced overlap and changes in the relative arrangement of knowledge items. These cases therefore reflect a broader reorganization of within-relation geometry, rather than only the displacement of an individual target item. This pattern illustrates the structural instability captured by KTS through changes in the ordering of pairwise representation distances. The MDS projections are provided for visualization only, whereas KTS is computed in the original representation space.

\subsection{More Quantitative Model Comparison}
\label{more_models}

The extended comparison across 43 additional configurations from 11 model families further shows that overall accuracy and representational stability do not necessarily improve together. For example, Phi-4 achieves higher ACC than Phi-2 (0.9240 vs.\ 0.8904) but lower LSS (0.5585 vs.\ 0.6770) and KTS (0.8033 vs.\ 0.8405); Mistral-Small-Instruct likewise achieves the highest ACC in its family (0.9549) but the lowest LSS (0.6725) and KTS (0.8310). Models with similar ACC can also differ substantially in stability: Gemma-2-9B and Yi-1.5-34B achieve comparable ACC (0.9614 vs.\ 0.9595), while Gemma-2-9B has higher LSS (0.7119 vs.\ 0.5849) and KTS (0.8478 vs.\ 0.8027). Although some model sequences exhibit joint improvements in correctness and stability, this pattern is not universal. These expanded comparisons support the conclusion that LSS and KTS capture dimensions of latent support that are not determined by model size or overall accuracy alone.

\begin{table*}[!t]
\centering
\caption{\textbf{Extended Model Rankings.}
ACC, LSS, and KTS are reported for 43 additional model configurations across 11 model families. The best and second-best results in each column are shown in bold and underlined, respectively.}
\label{tab:extended_model_rankings}
\setlength{\tabcolsep}{3pt}
\renewcommand{\arraystretch}{0.94}

\begin{tabular}{@{}
  >{\raggedright\arraybackslash}p{0.1\textwidth}
  >{\raggedright\arraybackslash}p{0.6\textwidth}
  >{\centering\arraybackslash}p{0.07\textwidth}
  >{\centering\arraybackslash}p{0.07\textwidth}
  >{\centering\arraybackslash}p{0.07\textwidth}@{}}
\toprule
Family & Model & ACC $\uparrow$ & LSS $\uparrow$ & KTS $\uparrow$ \\
\midrule

\multirow{5}{*}{\ModelIcon{gemma.png}\textit{Gemma}}
& \href{https://huggingface.co/google/gemma-7b}{\nolinkurl{google/gemma-7b}}
& 0.8939 & 0.6114 & \textbf{0.8643} \\

& \href{https://huggingface.co/google/gemma-1.1-2b-it}{\nolinkurl{google/gemma-1.1-2b-it}}
& 0.7936 & 0.6705 & 0.7747 \\

& \href{https://huggingface.co/google/gemma-1.1-7b-it}{\nolinkurl{google/gemma-1.1-7b-it}}
& 0.8580 & 0.6728 & 0.7932 \\

& \href{https://huggingface.co/google/gemma-2-2b}{\nolinkurl{google/gemma-2-2b}}
& 0.9198 & 0.7012 & 0.8397 \\

& \href{https://huggingface.co/google/gemma-2-9b}{\nolinkurl{google/gemma-2-9b}}
& \underline{0.9614} & \textbf{0.7119} & 0.8478 \\

\midrule

\multirow{3}{*}{\IconModel{Phi}\textit{Phi}}
& \href{https://huggingface.co/microsoft/phi-1_5}{\nolinkurl{microsoft/phi-1_5}}
& 0.6107 & 0.6258 & 0.7941 \\

& \href{https://huggingface.co/microsoft/phi-2}{\nolinkurl{microsoft/phi-2}}
& 0.8904 & 0.6770 & 0.8405 \\

& \href{https://huggingface.co/microsoft/phi-4}{\nolinkurl{microsoft/phi-4}}
& 0.9240 & 0.5585 & 0.8033 \\

\midrule

\multirow{3}{*}{\IconModel{Mistral}\textit{Mistral}}
& \href{https://huggingface.co/mistralai/Mistral-7B-v0.3}{\nolinkurl{mistralai/Mistral-7B-v0.3}}
& 0.9240 & \underline{0.7078} & 0.8556 \\

& \href{https://huggingface.co/mistralai/Mistral-Nemo-Base-2407}{\nolinkurl{mistralai/Mistral-Nemo-Base-2407}}
& 0.9321 & 0.6927 & 0.8511 \\

& \href{https://huggingface.co/mistralai/Mistral-Small-Instruct-2409}{\nolinkurl{mistralai/Mistral-Small-Instruct-2409}}
& 0.9549 & 0.6725 & 0.8310 \\

\midrule

\multirow{3}{*}{\IconModel{Granite}\textit{Granite}}
& \href{https://huggingface.co/ibm-granite/granite-7b-base}{\nolinkurl{ibm-granite/granite-7b-base}}
& 0.8067 & 0.6835 & 0.8429 \\

& \href{https://huggingface.co/ibm-granite/granite-3b-code-base-128k}{\nolinkurl{ibm-granite/granite-3b-code-base-128k}}
& 0.7207 & 0.6575 & 0.7978 \\

& \href{https://huggingface.co/ibm-granite/granite-8b-code-base-128k}{\nolinkurl{ibm-granite/granite-8b-code-base-128k}}
& 0.7396 & 0.6655 & 0.8264 \\

\midrule

\IconModel{Jamba}\textit{Jamba}
& \href{https://huggingface.co/ai21labs/AI21-Jamba2-3B}{\nolinkurl{ai21labs/AI21-Jamba2-3B}}
& 0.8800 & 0.6419 & 0.8319 \\

\midrule

\multirow{7}{*}{\IconModel{LFM}\textit{LFM}}
& \href{https://huggingface.co/LiquidAI/LFM2-350M}{\nolinkurl{LiquidAI/LFM2-350M}}
& 0.5756 & 0.5507 & 0.7754 \\

& \href{https://huggingface.co/LiquidAI/LFM2-700M}{\nolinkurl{LiquidAI/LFM2-700M}}
& 0.6617 & 0.6480 & 0.7946 \\

& \href{https://huggingface.co/LiquidAI/LFM2-1.2B}{\nolinkurl{LiquidAI/LFM2-1.2B}}
& 0.7122 & 0.6683 & 0.8082 \\

& \href{https://huggingface.co/LiquidAI/LFM2-2.6B}{\nolinkurl{LiquidAI/LFM2-2.6B}}
& 0.7535 & 0.6528 & 0.8220 \\

& \href{https://huggingface.co/LiquidAI/LFM2.5-230M}{\nolinkurl{LiquidAI/LFM2.5-230M}}
& 0.6566 & 0.5522 & 0.8032 \\

& \href{https://huggingface.co/LiquidAI/LFM2.5-350M}{\nolinkurl{LiquidAI/LFM2.5-350M}}
& 0.6671 & 0.5278 & 0.8009 \\

& \href{https://huggingface.co/LiquidAI/LFM2.5-1.2B-Base}{\nolinkurl{LiquidAI/LFM2.5-1.2B-Base}}
& 0.7508 & 0.6982 & 0.8275 \\

\midrule

\multirow{7}{*}{\IconModel{SmolLM}\textit{SmolLM}}
& \href{https://huggingface.co/HuggingFaceTB/SmolLM-135M}{\nolinkurl{HuggingFaceTB/SmolLM-135M}}
& 0.5093 & 0.5844 & 0.7946 \\

& \href{https://huggingface.co/HuggingFaceTB/SmolLM-360M}{\nolinkurl{HuggingFaceTB/SmolLM-360M}}
& 0.4923 & 0.6613 & 0.8111 \\

& \href{https://huggingface.co/HuggingFaceTB/SmolLM-1.7B}{\nolinkurl{HuggingFaceTB/SmolLM-1.7B}}
& 0.7006 & 0.6378 & 0.8303 \\

& \href{https://huggingface.co/HuggingFaceTB/SmolLM2-135M}{\nolinkurl{HuggingFaceTB/SmolLM2-135M}}
& 0.5910 & 0.6058 & 0.8104 \\

& \href{https://huggingface.co/HuggingFaceTB/SmolLM2-360M}{\nolinkurl{HuggingFaceTB/SmolLM2-360M}}
& 0.7400 & 0.6605 & 0.8316 \\

& \href{https://huggingface.co/HuggingFaceTB/SmolLM2-1.7B}{\nolinkurl{HuggingFaceTB/SmolLM2-1.7B}}
& 0.8634 & 0.6834 & 0.8470 \\

& \href{https://huggingface.co/HuggingFaceTB/SmolLM3-3B}{\nolinkurl{HuggingFaceTB/SmolLM3-3B}}
& 0.8684 & 0.7060 & 0.8402 \\

\midrule

\IconModel{ERNIE}\textit{ERNIE}
& \href{https://huggingface.co/baidu/ERNIE-4.5-0.3B-Base-PT}{\nolinkurl{baidu/ERNIE-4.5-0.3B-Base-PT}}
& 0.6570 & 0.6209 & 0.7793 \\

\midrule

\multirow{4}{*}{\IconModel{Hunyuan}\textit{Hunyuan}}
& \href{https://huggingface.co/tencent/Hunyuan-0.5B-Pretrain}{\nolinkurl{tencent/Hunyuan-0.5B-Pretrain}}
& 0.6142 & 0.5878 & 0.7652 \\

& \href{https://huggingface.co/tencent/Hunyuan-1.8B-Pretrain}{\nolinkurl{tencent/Hunyuan-1.8B-Pretrain}}
& 0.6605 & 0.6434 & 0.7991 \\

& \href{https://huggingface.co/tencent/Hunyuan-4B-Pretrain}{\nolinkurl{tencent/Hunyuan-4B-Pretrain}}
& 0.7932 & 0.6548 & 0.8274 \\

& \href{https://huggingface.co/tencent/Hunyuan-MT-7B}{\nolinkurl{tencent/Hunyuan-MT-7B}}
& 0.7041 & 0.6169 & 0.7905 \\

\midrule

\IconModel{MiMo}\textit{MiMo}
& \href{https://huggingface.co/XiaomiMiMo/MiMo-7B-Base}{\nolinkurl{XiaomiMiMo/MiMo-7B-Base}}
& 0.8349 & 0.6657 & \underline{0.8558} \\

\midrule

\multirow{8}{*}{\IconModel{Yi}\textit{Yi}}
& \href{https://huggingface.co/01-ai/Yi-6B}{\nolinkurl{01-ai/Yi-6B}}
& 0.7049 & 0.6132 & 0.8396 \\

& \href{https://huggingface.co/01-ai/Yi-9B}{\nolinkurl{01-ai/Yi-9B}}
& 0.9097 & 0.5556 & 0.8218 \\

& \href{https://huggingface.co/01-ai/Yi-34B}{\nolinkurl{01-ai/Yi-34B}}
& \textbf{0.9676} & 0.6306 & 0.8460 \\

& \href{https://huggingface.co/01-ai/Yi-1.5-6B}{\nolinkurl{01-ai/Yi-1.5-6B}}
& 0.9267 & 0.5589 & 0.8242 \\

& \href{https://huggingface.co/01-ai/Yi-1.5-9B}{\nolinkurl{01-ai/Yi-1.5-9B}}
& 0.9383 & 0.5579 & 0.8156 \\

& \href{https://huggingface.co/01-ai/Yi-1.5-34B}{\nolinkurl{01-ai/Yi-1.5-34B}}
& 0.9595 & 0.5849 & 0.8027 \\

& \href{https://huggingface.co/01-ai/Yi-Coder-1.5B}{\nolinkurl{01-ai/Yi-Coder-1.5B}}
& 0.5289 & 0.6248 & 0.7784 \\

& \href{https://huggingface.co/01-ai/Yi-Coder-9B}{\nolinkurl{01-ai/Yi-Coder-9B}}
& 0.6439 & 0.6142 & 0.8294 \\

\bottomrule
\end{tabular}

\vspace{-0.3cm}
\end{table*}

\section{Details of Application}
\label{app:application-details}

\noindent\textbf{LatentDPO Problem Setup.}
We instantiate the stability-aware policy training method for each
backbone--editor pair, where the backbones are GPT2-XL and LLaMA~2-7B
and the editors are ROME, MEMIT, and SAKE. The evaluation contains
$N=1{,}998$ retained CounterFact edit requests. For every request $i$,
the policy selects between two frozen editing actions:
\begin{equation}
a_i\in\{0,1\},\qquad
a_i=0:\text{Base},\qquad
a_i=1:\text{Base}+\text{gated correction}.
\label{eq:app-actions}
\end{equation}
Here, Base denotes the output of the corresponding editor. ROME is
evaluated one request at a time with the original weights restored
between requests, MEMIT edits the complete request batch jointly, and
SAKE uses its standard transport edit. Policy training does not further
modify the language-model backbone or the underlying editor.

LatentDPO is an offline binary-action selector over these two frozen
editing actions. The gated correction is applied only to requests for
which the learned policy selects action~1, rather than uniformly to all
requests. Policy training does not further modify the language-model
backbone or the underlying editor.

\noindent\textbf{Gated Activation Correction.}
Starting from the final-token hidden state $\widetilde h_i(p)$ produced
by the base editor, we fit a request-specific correction vector
$\delta_i^\star$ using 100 training-source prompts:
\begin{equation}
\delta_i^\star
=
\arg\min_{\|\delta\|_2\le r_i}
\frac{1}{100}
\sum_{m=1}^{100}
\operatorname{CE}\!\left(
W_{\mathrm{out}}
\bigl(\widetilde h_i(p_{im})+\delta\bigr),
y_i
\right),
\label{eq:app-correction}
\end{equation}
where $y_i$ is the first token of the requested target and $r_i$ is the
median distance from the target-token activations to their mean. We
optimize Eq.~\eqref{eq:app-correction} for 40 Adam steps with learning
rate $r_i/(10\sqrt{d})$ and project the correction onto the
$\ell_2$ ball after every step.

When leading-space and no-leading-space tokenizations produce different
target-token IDs, we select the token according to its target-cloud
exact rate and then its mean target logit. A frozen
\texttt{paraphrase-MiniLM-L6-v2} encoder provides prompt embeddings.
Let $\phi(p)$ denote the embedding of prompt $p$, and let $\mu_i$ be
the mean embedding of the 100 training-source prompts and the canonical
rewrite prompt. The scope gate and corrected state are
\begin{align}
g_i(p)
&=
\mathbf{1}\!\left[
\|\phi(p)-\mu_i\|_2\le\tau_{b,e}
\right],\\
h_i^{(1)}(p)
&=
\widetilde h_i(p)
+\alpha_{b,e}g_i(p)\delta_i^\star,
\label{eq:app-gated-action}
\end{align}
where $\alpha_{b,e}$ and $\tau_{b,e}$ denote the correction strength
and gate threshold for backbone $b$ and editor $e$. This correction is
applied to activations at inference time and does not modify the model
weights or output matrix.

For ROME and MEMIT, the correction configurations are selected
conservatively to limit the observed Specificity loss while retaining
a nontrivial efficacy gain. SAKE retains the correction configuration
used in its original policy experiment. These operating points are
fixed before policy fitting:

\begin{table}[t]
\centering
\small
\caption{\textbf{Frozen Gated-Correction Configurations.}
Correction strength $\alpha_{b,e}$ and prompt-gate threshold
$\tau_{b,e}$ used to construct action $1$.}
\label{tab:app-action-configurations}

\begin{tabular}{clcc}
\toprule
Backbone & Editor & $\alpha_{b,e}$ & $\tau_{b,e}$ \\
\midrule

\multirow{3}{*}{GPT2-XL}
& ROME  & 0.75 & 5.25 \\
& MEMIT & 0.50 & 5.50 \\
& SAKE  & 1.00 & 6.75 \\

\midrule

\multirow{3}{*}{LLaMA~2-7B}
& ROME  & 0.75 & 5.25 \\
& MEMIT & 0.25 & 5.55 \\
& SAKE  & 1.00 & 6.75 \\

\bottomrule
\end{tabular}

\vspace{-0.5cm}
\end{table}

\noindent\textbf{Application-Specific Stability Features.}
The available editing cache contains final-layer activation clouds for
three distributions:
\[
d\in\{\text{source},\text{forced source},\text{target}\}.
\]
Because these caches do not contain the five elicitation conditions,
multiple J-Lens layers, and external development data required by the
full LSS/KTS protocol, we construct lightweight application-specific
proxies derived from the same item-identity and relation-geometry
principles.

For each request and distribution, the 100 training activations are
randomly partitioned into five fixed groups of 20. After row
normalization, each group is represented by its normalized centroid,
projected to 64 dimensions using a fixed Gaussian random projection,
and normalized again. The proxy-extraction seed is $20{,}260{,}830$.
We denote the resulting five views by
$u_{i,d}^{(0)},\ldots,u_{i,d}^{(4)}$.

For the LSS-inspired feature, facts are divided into five
within-relation cross-fitting folds. For every held-out fold $f$,
distribution $d$, and view $v>0$, a similarity-Procrustes map
$P_{d,v}^{(-f)}$ is fitted on the other four folds to align view $v$
with reference view 0. Let $s_{r_i}^{(-f)}$ denote the median pairwise
distance between reference-view centroids for the training facts in
relation $r_i$, with the global training median used when a
relation-specific estimate is unavailable. We define
\begin{align}
D_{i,d}
&=
\operatorname*{median}_{v=1,\ldots,4}
\frac{
\left\|
P_{d,v}^{(-f(i))}
\bigl(u_{i,d}^{(v)}\bigr)-u_{i,d}^{(0)}
\right\|_2
}{
s_{r_i}^{(-f(i))}
},\\
\widehat{\mathrm{LSS}}_{i,d}
&=\frac{1}{1+D_{i,d}},
\qquad
\widehat{\mathrm{LSS}}_i
=\frac{1}{3}\sum_d\widehat{\mathrm{LSS}}_{i,d}.
\label{eq:app-partial-iss}
\end{align}

For the KTS-inspired feature, the five views are centered and normalized
separately within each relation and activation distribution. Across
the ten view pairs, we measure bidirectional identity retrieval
adjusted for its $1/n_r$ chance rate, Spearman preservation of
within-relation cosine-distance structure, and overlap between the
three-nearest-neighbor sets. The primary composite is
\begin{equation}
\widehat{\mathrm{KTS}}_{i,d}
=
\frac{2I_{i,d}G_{r_i,d}}
{I_{i,d}+G_{r_i,d}+10^{-12}},
\qquad
\widehat{\mathrm{KTS}}_i
=
\frac{1}{3}\sum_d\widehat{\mathrm{KTS}}_{i,d},
\label{eq:app-partial-kts}
\end{equation}
where $I_{i,d}$ is chance-adjusted identity retrieval and $G_{r_i,d}$
is normalized geometry preservation. The three-nearest-neighbor
overlap is retained as an additional policy feature but is not included
in Eq.~\eqref{eq:app-partial-kts}.

Within every training partition, the stability features are converted
to empirical percentiles using only the corresponding training facts:
\begin{equation}
q_i^{\mathrm{LSS}}
=
\widehat F_{\mathrm{LSS}}^{\mathrm{train}}
\bigl(\widehat{\mathrm{LSS}}_i\bigr),
\qquad
q_i^{\mathrm{KTS}}
=
\widehat F_{\mathrm{KTS}}^{\mathrm{train}}
\bigl(\widehat{\mathrm{KTS}}_i\bigr).
\label{eq:app-stability-percentiles}
\end{equation}
We use midpoint ranks for ties and assign a value of $0.5$ when a
feature is unavailable.

\noindent\textbf{Preference Construction.}
Let $A_{i,a}\in\{0,1\}$ indicate whether the greedy continuation from
the canonical rewrite prompt begins with the requested target token
under action $a$. We construct a 70/30 preference utility:
\begin{align}
R_{i,0}
&=
0.70A_{i,0}
+0.15(1-q_i^{\mathrm{LSS}})
+0.15(1-q_i^{\mathrm{KTS}}),\\
R_{i,1}
&=
0.70A_{i,1}
+0.15q_i^{\mathrm{LSS}}
+0.15q_i^{\mathrm{KTS}}.
\label{eq:app-preference-reward}
\end{align}
Canonical editing success therefore receives 70\% of the utility,
whereas the item- and relation-level stability features receive 15\%
each. Generality and Specificity are not included directly in the
training utility. For ROME and MEMIT, locality is instead controlled
through the conservative action configurations in
Table~\ref{tab:app-action-configurations}. The preferred action is
$a_i^+=1$ when $R_{i,1}\ge R_{i,0}$ and $a_i^+=0$ otherwise.

\noindent\textbf{Policy Inputs and Optimization.}
Each request is represented by 14 features. Six describe the fitted
correction: mapped-source success and cross-entropy, target success and
cross-entropy, trust radius, and unscaled correction norm. Eight
describe the partial geometry: source, target, and mean LSS; source
identity retrieval, three-nearest-neighbor overlap, and composite KTS;
mean KTS; and the geometric mean of the global LSS and KTS ranks.
Missing values are replaced by training-partition medians, and all
features are standardized using training-partition statistics.

The editing-action policy is a linear Bernoulli model:
\begin{equation}
\pi_\theta(1\mid x_i)
=
\sigma(w^\top z_i+c),
\qquad
\pi_\theta(0\mid x_i)
=
1-\pi_\theta(1\mid x_i),
\label{eq:app-policy}
\end{equation}
where $z_i$ is the standardized feature vector. With a uniform
reference policy, we optimize
\begin{equation}
\mathcal{L}_{\mathrm{pref}}
=
-\frac{1}{|\mathcal{D}_{\mathrm{train}}|}
\sum_i
\log\sigma\!\left(
\beta
\left[
\log\frac{\pi_\theta(a_i^+\mid x_i)}
{\pi_{\mathrm{ref}}(a_i^+\mid x_i)}
-
\log\frac{\pi_\theta(a_i^-\mid x_i)}
{\pi_{\mathrm{ref}}(a_i^-\mid x_i)}
\right]
\right)
+
\frac{\lambda}{2}\|w\|_2^2,
\label{eq:app-policy-objective}
\end{equation}
where $a_i^-=1-a_i^+$. We optimize the policy for 1,500 Adam steps
with learning rate $0.03$. We select
\[
\beta\in\{0.1,0.3,1,3\},
\qquad
\lambda\in\{0,10^{-4},10^{-3},10^{-2},10^{-1}\}
\]
using relation-grouped validation reward. At inference time, action 1
is selected when $\pi_\theta(1\mid x_i)\ge0.5$.

\noindent\textbf{Relation-Disjoint Evaluation.}
The 34 CounterFact relations are assigned to five row-balanced outer
folds containing 407, 402, 402, 392, and 395 requests. No relation
appears in both the training and test partitions of an outer fold.
Hyperparameters are selected using an inner relation-grouped validation
split. The selected policy is then refitted on the complete outer
training partition and applied once to the held-out relations.
Concatenating the five test partitions produces exactly one
out-of-fold decision for every request. The split seed is
$20{,}260{,}915$.

Accuracy is exact success on the canonical rewrite prompt. Generality
is mean success over two held-out CounterFact paraphrases. Specificity
is the edit-macro average over eligible neighborhood prompts, excluding
prompts for which the unedited model already prefers the new target.
All metrics are reported over the same 1,998 requests.

\noindent\textbf{Scope of Application.}
The experiment evaluates an offline policy that selects between two
previously fitted editing actions. The action library is fixed before
policy fitting, and every reported policy decision is obtained out of
fold with respect to relation. The application-specific stability
features are lightweight proxies derived from LSS and KTS rather than
the formal multilayer benchmark metrics.

\section{Theoretical Derivation}
\label{app:theoretical-derivation}

\subsection{Representation Extraction and Cross-fitting}
\label{app:representation-crossfitting}

For every query, we extract the residual-stream state at the final non-padding
input token after the complete question has been processed and before the first
answer token is generated. Layer depth is normalized as
$\lambda_\ell=\ell/(L-1)$, and the primary scores use the final-quarter window
$\mathcal W_m=\{\ell:\lambda_\ell\geq0.75\}$. J-Lens estimates a rank-256 transport
from each selected layer to the final representation space using randomized
range finding. The transport is calibrated separately for each model and layer
on 128 length-128 sequences sampled from \texttt{NeelNanda/pile-10k}, and the
same fitted transport is then applied to all items and access conditions.

The similarity alignment used by LSS is estimated with five-fold cross-fitting.
Items are assigned to folds within each track while preserving the distribution
of comparison groups, and all queries associated with an item remain in the
same fold. For an item in fold $f$, the condition-to-reference map is fitted using
the other four folds and then applied to the held-out item. The local reference
scale $\beta^\ell_{m,r_i}$ is estimated from the corresponding reference
partition. Cross-fitting therefore prevents the alignment or normalization
scale from being chosen to make the evaluated item appear stable.

\subsection{Aggregation And Statistical Levels}
\label{app:aggregation-levels}

For LSS, the normalized residual $d^\ell_{m,i,t}$ is first summarized by the
median over the available non-reference elicitation conditions and then averaged over
$\mathcal W_m$. We denote this aggregated drift by
\begin{equation}
 D_{m,i}=\frac{1}{|\mathcal W_m|}\sum_{\ell\in\mathcal W_m}
 \underset{t\in\mathcal T_i}{\operatorname{median}}
 d^\ell_{m,i,t}.
 \label{eq:app-iss-aggregation}
\end{equation}
Statistical comparisons use $D_{m,i}$, while
$\operatorname{LSS}_{m,i}=(1+D_{m,i})^{-1}$ is reported as its bounded,
monotone transformation. Item-level LSS values are averaged within each track;
the two track means receive equal weight in an overall benchmark score.

KTS is defined at the comparison-group level. For each condition pair
$p=(t,t')$, centering uses exactly the items observed under both conditions, so
the two distance matrices have identical membership. A condition pair is valid
only when the shared set meets the pre-specified size requirement and both
distance vectors admit a Spearman correlation. Correlations are averaged
equally across valid condition pairs and selected layers to obtain
$\operatorname{KTS}_{m,r}$. Group scores are then macro-averaged within each
track, followed by an equal-weight average of the two tracks. When KTS is used
with an item-level outcome, item $i$ is assigned the group-level context
$\operatorname{KTS}_{m,r_i}$; this assignment does not make KTS an item-level
measurement.

\subsection{Properties Of the Stability Measures}
\label{app:metric-properties}

Correctness does not uniquely determine the internal state that produces an
answer. For a single generated token with output weights $w_j$ and biases
$b_j$, predicting token $y$ requires
\begin{equation}
 (w_y-w_j)^\top u\geq b_j-b_y,\qquad \forall j\neq y.
 \label{eq:app-output-region}
\end{equation}
These inequalities define a region of representations that produce the same
token. Distinct points in this region can therefore yield the same output while
having different locations and relations to other items. For multi-token
answers, the same argument applies to the sequence of output-compatible regions
visited during generation.

LSS is invariant to a condition-wide change that is exactly captured by the
fitted similarity map. If
$A^\ell_{m,\tau,t}(\bar v^\ell_{m,i,t})=\bar v^\ell_{m,i,a}$ for every item,
then $e^\ell_{m,i,t}=0$, $D_{m,i}=0$, and
$\operatorname{LSS}_{m,i}=1$. More generally,
\begin{equation}
 \frac{\partial\operatorname{LSS}_{m,i}}{\partial D_{m,i}}
 =-(1+D_{m,i})^{-2}<0,
 \label{eq:app-iss-monotonicity}
\end{equation}
so LSS decreases monotonically with normalized residual drift.

For KTS, group centering removes translations shared by all items in a
comparison group. Pairwise cosine distance is unchanged when the centered
representations undergo a common positive isotropic scaling and orthogonal
transformation. Spearman correlation further remains unchanged under a strictly
monotone transformation of all pairwise distances. Consequently,
$\operatorname{KTS}_{m,r}=1$ whenever the ordering of within-group pairwise
distances is perfectly preserved across the compared conditions, even if their
absolute values change.

\subsection{Interpretation And Scope}
\label{sec:metric-interpretation}

LSS is the normalized residual displacement of an individual item's
condition-level representation, whereas KTS is the distance-rank preservation
of its comparison group's internal organization. Neither metric contains a
correctness criterion: stable representations may support an incorrect target,
and correct outputs may arise from representations that vary across access
conditions. The two measures are therefore geometric diagnostics of latent
support and complement behavioral correctness.

Because the observed state precedes the first generated answer token, the
metrics characterize pre-generation support. They do not measure the stability
of the full hidden-state trajectory produced during multi-token generation.

\section{Limitation}
\label{lim}
This work primarily focuses on evaluating internal knowledge stability rather than directly optimizing it. In future work, we plan to build on the principles underlying LSS and KTS to develop post-training strategies that explicitly improve the stability of internal knowledge representations. These metrics may serve as training objectives, regularization signals, or feedback criteria to encourage consistent knowledge access across different expressions, contexts, formats, and languages. We will further investigate whether improving internal stability can enhance the reliability and generalization of model behavior while preserving overall task performance.

\section{More Discussions}
\label{dis}

\noindent$\triangleright$ \textbf{\textit{Q1. What do LSS and KTS measure, and does high stability necessarily imply correct knowledge?}}

LSS and KTS should be interpreted as complementary to correctness rather than as substitutes for it. LSS measures whether an individual knowledge item preserves its representational identity after condition-wide shifts are removed, whereas KTS measures whether the organization among items within the same relation remains consistent. A model may stably represent an incorrect association; therefore, stability alone does not guarantee factual correctness. Conversely, a correct response may be supported by an unstable representation and fail under another form of access. We consequently advocate jointly examining ACC, LSS, and KTS: ACC captures answer correctness, LSS captures item-level stability, and KTS captures structural coherence. Their combination distinguishes reliably supported correct answers from correct but internally fragile behavior.

\noindent$\triangleright$ \textbf{\textit{Q2. Could LSS and KTS mainly reflect differences in wording, format, or language?}}

Several results suggest that the proposed metrics are not dominated by superficial prompt variation. LSS first removes representational shifts shared across items through relation-stratified, cross-fitted alignment, retaining only item-specific residual drift. Although this residual accounts for only 4.2\%–6.8\% of the raw cross-condition drift, it provides an additional 1.20 AUROC points for predicting answer changes, whereas the wrong-condition control produces only a near-zero gain. The semantic perturbation experiment further reveals a graded response: minimal edits induce the smallest changes, controlled probes produce intermediate changes, and same-relation hard negatives produce the largest changes. The stable rankings under leave-one-probe-out evaluation provide additional evidence that the results are not driven by any single probe type.

\noindent$\triangleright$ \textbf{\textit{Q3. Why are the final-quarter layers and J-Lens representations used?}}

The final-quarter window is motivated by both sensitivity analysis and causal intervention. Representations from the 80\%–100\% depth interval almost reproduce the primary LSS and KTS rankings, whereas earlier intervals show substantially weaker agreement. More importantly, activation patching shows that late-layer states contain increasingly behaviorally effective item-specific information: the recovery advantage of same-item over same-relation control patches increases from +0.02 in the 40\%–50\% interval to +0.80 in the final quarter. J-Lens provides a shared comparison space across layers, while aggregation over multiple late layers reduces dependence on a single state. Thus, the selected representation is not intended to describe all stages of computation; it operationalizes stability in the late, pre-generation states most directly involved in producing the answer.

\noindent$\triangleright$ \textbf{\textit{Q4. Why are both LSS and KTS necessary?}}

LSS and KTS capture different levels of latent organization. LSS is defined at the knowledge-item level and measures whether a particular item remains distinguishable across elicitation conditions. KTS is defined at the relation or task-type level and measures whether the relative geometry among multiple items is preserved. These properties need not change together: an individual item may drift while the broader relational structure remains stable, or several moderate item-level shifts may collectively reorganize the relation geometry. This distinction is also reflected in the non-parallel model and condition rankings observed in our experiments. Accordingly, KTS is treated as a relation-level contextual property rather than an item-specific score. Using both metrics provides a more complete account of latent support than either item identity or relational structure alone.

\noindent$\triangleright$ \textbf{\textit{Q5. Can consistently incorrect knowledge also receive high stability scores?}}

Yes, consistently incorrect knowledge can also receive high stability scores. This is why we evaluate LSS and KTS jointly with ACC: stability complements correctness rather than replacing it.

\end{document}